\documentclass[journal,twoside,web]{ieeecolor}
\usepackage{tmi}
\usepackage{cite}
\usepackage{amsmath,amssymb,amsfonts}
\usepackage{algorithmic}
\usepackage{graphicx}
\usepackage{textcomp}

\usepackage{booktabs}
\usepackage{multirow}
\usepackage{url}
\usepackage{subfig}
\usepackage{algorithm}

\usepackage{ulem}
\usepackage{cancel}

\newcommand{\etc}{ \textit{etc.} }
\newcommand{\eg}{ \textit{e.g.}, }
\newcommand{\ie}{ \textit{i.e.}, }

\def\BibTeX{{\rm B\kern-.05em{\sc i\kern-.025em b}\kern-.08em
    T\kern-.1667em\lower.7ex\hbox{E}\kern-.125emX}}
\begin{document}
\title{ImageGCN: Multi-Relational Image Graph Convolutional Networks for Disease Identification with Chest X-rays}
\author{Chengsheng Mao, Liang Yao, and Yuan Luo

\thanks{The research is supported in part by the following US NIH grants: R21LM012618, 5UL1TR001422, U01TR003528 and R01LM013337.}
\thanks{C. Mao, L. Yao and Y. Luo are with Northwestern University,  Chicago IL 60611, United States.
(e-mail: \{chengsheng.mao, liang.yao, yuan.luo\}@northwestern.edu)}
}

\maketitle

\begin{abstract}
Image representation is a fundamental task in computer vision. However, most of the existing approaches for image representation ignore the relations between images and consider each input image independently. Intuitively, relations between images can help to understand the images and maintain model consistency over related images, leading to better explainability. In this paper, we consider modeling the image-level relations to generate more informative image representations, and propose ImageGCN, an end-to-end graph convolutional network framework for inductive multi-relational image modeling. We apply ImageGCN to chest X-ray images where rich relational information is available for disease identification. Unlike previous image representation models, ImageGCN learns the representation of an image using both its original pixel features and its relationship with other images. Besides learning informative representations for images, ImageGCN can also be used for object detection in a weakly supervised manner. The experimental results on 3 open-source x-ray datasets, ChestX-ray14, CheXpert and MIMIC-CXR demonstrate that ImageGCN can outperform respective baselines in both disease identification and localization tasks and can achieve comparable and often better results than the state-of-the-art methods.
\end{abstract}


\section{Introduction}
\label{sec:introduction}
\IEEEPARstart{L}{earning} low-dimensional representation of images is a fundamental task in computer vision. Deep learning techniques, especially the convolutional neural network (CNN) architectures have achieved remarkable breakthroughs in learning image representation for classification \cite{krizhevsky2012imagenet,he2016deep,huang2017densely}. However, most of the existing approaches for image representation only considered each input image independently while ignored the relations between images. In reality, multiple relations can exist between images, especially in clinical setting, \eg medical images from the same person can show pathophysiologic progressions.  Intuitively, related images can give certain insights to better understand the current image. For example, images present in the same web page can help to understand each other; knowing a patient's other medical images can help to analyze the current image.

\begin{figure*}
  \centering
  \includegraphics[width=.98\linewidth,page=1]{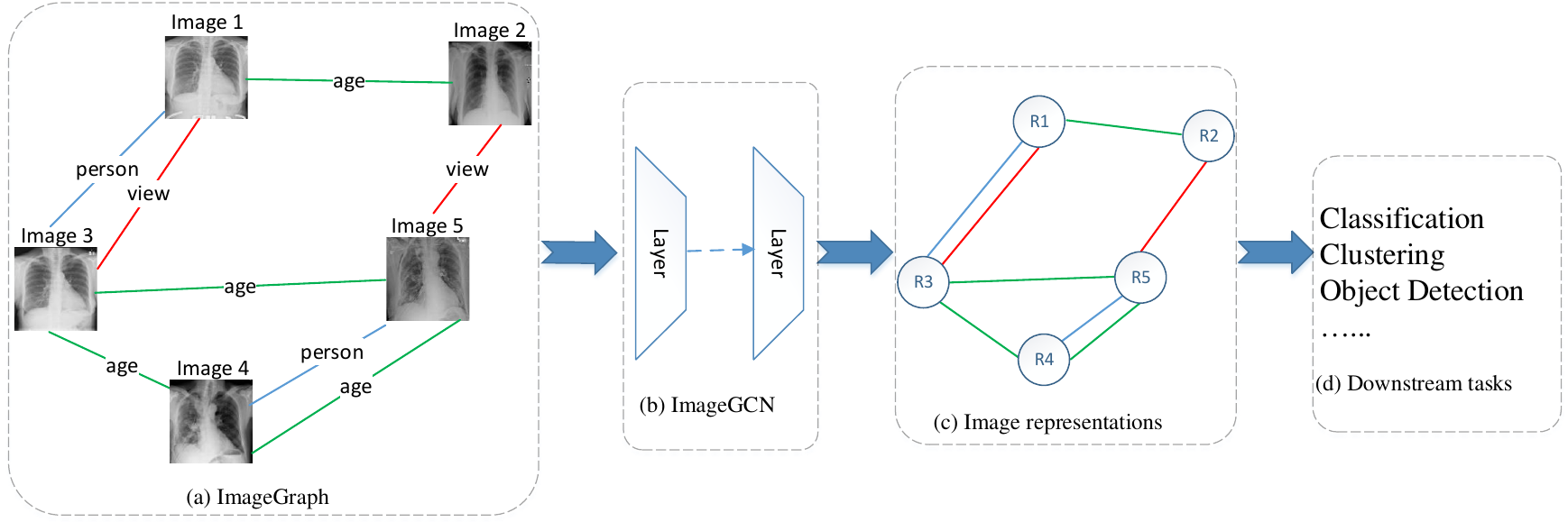}
  \caption{Overview of ImageGCN.
  (a) The ImageGraph constructed with the original images and the relations between them. Here we show 3 types of relations in an ImageGraph of CXR marked with different colors. The relations between CXR images are defined in Section \ref{sec:graphdefine}.
  (b) Multi-layers ImageGCN to model the ImageGraph. (c) The output low-dimensional distributed representations for all images in the ImageGraph. The structure of the graph is preserved. (d) The image representations are used for downstream tasks, such as classification, clustering, \etc
  }
  \label{fig:ImageGraph}
\end{figure*}

We model the images and the relations between them as a graph, named ImageGraph, where a node corresponds to an image and an edge between two nodes represents a relation between the two corresponding images. An ImageGraph incorporating multiple types of relations is a multigraph where multiple edges exist between two nodes. The neighborhood of an image in the ImageGraph represents the images that have close relations with it.
Fig. \ref{fig:ImageGraph}(a) shows an example of ImageGraph of chest X-ray (CXR) images incorporating 3 types of relations between 5 nodes.

Learning an image representation incorporating both neighborhood information and the original pixel information is not trivial, because the neighborhood information is unstructured and varies for different nodes.
Inspired by the emerging research on graph convolutional networks (GCN) \cite{kipf2017semi,hamilton2017inductive,chen2018fastgcn,schlichtkrull2018modeling} that can model graph data to learn informative representations for nodes based on the original node features and the structure information, we propose ImageGCN, an end-to-end GCN framework on ImageGraph, to learn the image representations. In ImageGCN, each image updates the information based on its own features and the images that have certain relation with it. Fig. \ref{fig:ImageGraph} shows an overview of ImageGCN, where each node in an ImageGraph is transformed into an informative representation by a number of layers. {In Fig. \ref{fig:ImageGraph}, the proposed model ImageGCN (b) accepts the input of ImageGraph (a) and outputs the representation of images (c) that fed to the down stream tasks (d).}

There are several issues when applying the original GCN \cite{kipf2017semi} to an ImageGraph. (1) The original GCN is transductive and requires all node features present during training, which does not scale out to large ImageGraphs. (2) The original GCN is for simple graphs and can not support the multi-relational ImageGraphs. (3) The original GCN is effective for low-dimensional feature vectors in nodes, and cannot be effectively extended to nodes with high-dimensional or unstructured features in ImageGraphs. Thanks to GraphSAGE \cite{hamilton2017inductive}, the inductive learning issue was addressed for GCN; the multi-relational issue was also addressed by relational GCN (R-GCN) \cite{schlichtkrull2018modeling}. However, applying GCN to high-dimensional or unstructured features still remains unaddressed. The proposed ImageGCN in this paper is able to address this issue by designing appropriate message passing units.

In ImageGCN, we propose to design flexible message passing units (MPU) to do message passing between two adjacent image nodes, instead of a linear transformation in the original GCN for flat feature vectors. We use a number of MPUs equipped with a multi-layer CNN architecture for message passing between images in a multi-relational ImageGraph. We introduce partial parameter sharing between different MPUs corresponding to different relations to reduce model complexity. We also incorporate the idea of GraphSAGE \cite{hamilton2017inductive} and R-GCN \cite{schlichtkrull2018modeling} to our ImageGCN model for inductive batch propagation on multi-relational ImageGraphs.

We evaluate ImageGCN on 3 open-source x-ray datasets, ChestX-ray14 \cite{wang2017chestxray}, CheXpert \cite{irvin2019chexpert} and MIMIC-CXR \cite{johnson2019mimic},  where rich relations are available between the Chest X-ray (CXR) images. The experimental results demonstrate that ImageGCN can outperform the respective baselines in both disease identification and localization. The source code is available at \url{https://github.com/mocherson/ImageGCN}.

In summary, the main contributions are as follows. (1) To our best knowledge, this is the first study to model natural image-level relations for image representation. (2) from the perspective of method, we propose ImageGCN to extend GCN to graphs of high-dimensional and unstructured node features using a flexible message passing unit (MPU) to do message passing between two adjacent notes, instead of a linear transformation in the original GCN for flat feature vectors; (3) in our framework, to reduce the complexity of ImageGCN, we introduce the partial parameter sharing scheme (PPS) among the MPUs of GCN; (4) we also incorporate into ImageGCN the idea of R-GCN for multi-relational graphs, and GraphSAGE for inductive batch propagation; (5) from the perspective of application, we applied our framework to an ImageGraph of chest X-ray and improved the performance of both disease identification and localization.

\section{Related work}
\subsection{Deep learning with CXR}
Since the ChestX-ray14 dataset \cite{wang2017chestxray} was released, an increasing amount of research on CXR image analysis have used deep neural networks for disease identification \cite{wang2017chestxray,yao2017learning,kumar2018boosted,mao2018deep,guan2018multi,yan2018weakly,chen2019deep,allaouzi2019novel,guendel2018learning,baltruschat2019comparison}. The general idea of previous studies is to generate a low-dimensional image representation by a deep neural network architecture, treating each image independently and ignoring the relations between images. They mainly focused on the design of the network architecture. Different from previous studies, in our work, besieds the original CXR images, we also involve the natural relations between CXR images into our framework to learn the image representation.

The previous research on relational model for image analysis mainly focused on pixel-level relations \cite{maire2016affinity,monti2017geometric}, object-level relations \cite{yao2010grouplet,marino2017more,dai2017detecting,zhu2018deep,yao2018exploring} and label-level relations \cite{lee2018multi,wang2011image,chen2019multi}. Similarity relations between images were also studied in literature \cite{garcia2017few,wang2011image}. Liu et al. \cite{liu2020semi} proposed a framework to preserve the similarity relation between image representations. However, the similarity relation cannot be regarded as a natural relation between images, because this relation can be directedly derived from the image representations. Few studies in literature are found to model the natural image-level relations for image representation. In this paper, we propose to employ the graph convolutional network to model the natural relations between images.

\subsection{Graph Convolutional Networks}
Since Graph Convolutional Networks (GCN) \cite{kipf2017semi} successfully generalized the operation of convolution to graphs and generated more informative representations for nodes and edges to improve the scalability and classification performance in large-scale graphs, a number of GCN variants have been developed to improve the GCN model at various aspects. GraphSAGE \cite{hamilton2017inductive} presented a general inductive graph convolutional network to efficiently generate node embeddings for previously unseen data. GraphSAGE also introduced a batch propagation algorithm by sampling and aggregating features from a node’s local neighborhood for training on large graphs. FastGCN \cite{chen2018fastgcn} not only improved GCN for inductive learning and batch propagation, but also reduced gradient computation to speed up the training process by importance sampling. Schlichtkrull et al. \cite{schlichtkrull2018modeling} introduced R-GCN to encode multiple types of relations into GCN for link prediction and entity classification. However, few studies in literature are found to extend GCN to a graph with high-dimensional or unstructured node features, \eg graph with images as nodes. In this work, we are making this extension and incorporate the merits of GraphSAGE and R-GCN in our framework.

\subsection{GCN for Image Classification}
Recently, GCN is increasingly explored for image classification. Wang et al. \cite{wang2018zero} implemented GCN on knowledge graphs to incorporate both semantic embeddings of class names and the relationships between classes for zero-shot image recognition. Chen et al. \cite{chen2019multi} proposed a multi-label GCN that builds a directed graph over the object labels represented by word embeddings to capture the label correlations for multi-label image recognition. Garcia and Bruna \cite{garcia2017few} proposed to build a fully-connected graph for all images based on the similarity, and leverage the graph neural network for semi-supervised few-shot learning. Hong et al. \cite{hong2020graph} developed miniGCN over a graph of hyperspectral pixels for hyperspectral image classification. Yu et al. \cite{yu2021cgnet,yu2021resgnet} proposed ResGNet-C and CGNet that first apply a CNN architecture to extract features of image and then apply a GCN over the k-nearest neighbor graph to generate image representations for disease detection. {Huang et al. \cite{huang2020edge} proposed an edge-variational GCN framework to integrate multimodal data for disease prediction with uncertainty estimation.} Different from previous studies, in this paper, we take into account the natural image-level relations to construct a multi-relational ImageGraph, and use ImageGCN to model the relations to learn informative representations for the nodes \ie images.

In the study by Chen et al. \cite{chen2019multi}, the graph was constructed with labels as nodes and label correlation dependencies as edges. This is a small graph with size equal to the number of labels. GCN was applied to this graph to get the label representations combined with an image representation for multi-label classification. In the study by Hong et al. \cite{hong2020graph}, the graph nodes are hyperspectral image pixels, and the edge between two nodes represents their similarity. In the studies by Yu et al. \cite{yu2021resgnet,yu2021cgnet}, a node in the graph is a {low-dimensional} image representation that is generated from a pre-trained and fine-tuned CNN architecture, each node is connected to its k nearest neighbors by Euclidean distance of image representations. This is a two-step training framework rather than an end-to-end framework, because they first train the CNN architecture to get the initial image representations, and then train a GCN to get the final image representations for classification. In addition, the similarity {between image representations} cannot be regarded as a natural relation between images, because this relation can be directly derived from the image representations. {In the study by Huang et al. \cite{huang2020edge}, the node features are also low-dimensional image representations from a pretrained feature extractor, but the edge weight between two nodes are learned from an encoder based on each pair of the non-imaging data of the two nodes. Tariq et al. \cite{tariq2021fusion} used a similar graph construction method for disease trajectory prediction, where the nodes are also image representations from a pretrained CNN architecture, and the edges are based on the similarity between patients' demographic information or medical history.} In our work, a graph node is an original image {represented by pixels}, and the edge represents the natural relations between images, our framework also supports multiple types of image relations (age relation, person relation, \etc), thus our graph is a multi-relational graph.

\section{Methods}
To make a clear understanding of our methods, we list the definition of common notations and symbols used throughout the paper in Table \ref{tab:notation}.

\begin{table}
\centering
\caption{Common notations used throughout the paper.}
\begin{tabular}{lp{65mm}}
\toprule
Notation  & Definition \\
\midrule
$H^{(k)}$       & a matrix of all the node representation at the $k$th layer     \\
$h^{(k)}_i$       & the representation of node $i$ at the $k$th layer     \\
$A$       & a djacency matrix of a graph     \\
$R$ & a set of relation types in the graph. in our work 4 types relations are considered.   \\
$A_r$       & normalized adjacency matrix for relation $r$ in a graph   \\
$a_{ij}$ ($a_{ij}^r)$  & the $i$th row and $j$th column in $A$ ($A_r$  ) \\
$W^{(k)}$       & the trainable linear transformation matrix at the $k$th layer        \\
$ W^{(k)}_r$       & the trainable linear transformation matrix for relation $r$ at the $k$th layer        \\
$\phi(\cdot)$ &    the activation function (ReLU in this paper)  \\
$N(i)$ & the set of all nodes that have a connection with node $i$ (self included) \\
$N_r(i)$ & the set of all nodes that have a connection with node $i$ with regards to relation $r$ (self included) \\
$f_r^{(k)}$ &  the kernel Message Passing Unit corresponding to relation $r$ in layer $k$, it is a function with learnable parameters.   \\
$f^{(k)}$ &  the sharing part of MPU by all relations in layer $k$   \\
$g_r^{(k)}$ &  the private part of MPU for relation $r$ in layer $k$   \\
\bottomrule
\end{tabular}
\label{tab:notation}
\end{table}

\subsection{Graph Convolutional Networks}
Graph convolutional network (GCN) \cite{kipf2017semi} can  incorporate the node feature information and the structure information to learn informative representations for nodes in the graph. GCN learns node representations with the following propagation rule derived from spectral graph convolutions for an undirected graph \cite{kipf2017semi}:
\begin{equation}\label{eq:GCN}
H^{(k+1)} = \phi (\tilde{D}^{-\frac{1}{2}} \tilde{A} \tilde{D}^{-\frac{1}{2}} H^{(k)} W^{(k)})
\end{equation}
where $\tilde{A}=A+I$ is the adjacency matrix with added self-connection, $\tilde{D}$ is a diagonal matrix with $\tilde{D}_{ii}=\sum_j{\tilde{A}_{ij}}$, $\tilde{D}^{-\frac{1}{2}} \tilde{A} \tilde{D}^{-\frac{1}{2}}$ can be seen as a symmetrically normalized adjacency matrix,  $H^{(k)}$ and $W^{(k)}$ are the node representation matrix and the trainable linear transformation matrix in the $k$th layer, $H^{(0)}$ is the original feature matrix of nodes, $\phi(\cdot)$ is the activation function (such as the ReLU).

The propagation rule of GCN in Eq. \ref{eq:GCN} can be interpreted as the Laplacian smoothing for a graph \cite{li2018deeper}, \ie the new feature of a node is computed as the weighted average of itself and its neighbors, followed by a linear transformation before activation function, \ie Eq. \ref{eq:nodegcn},
\begin{equation}\label{eq:nodegcn}
h^{(k+1)}_i = \phi \left(\sum_{j\in N(i)} c_{ij} h_j^{(k)}W^{(k)} \right)
\end{equation}
where $h_i^{(k)}$ is the representation of node $i$ in the $k$th layer, $N(i)$ is the set of all nodes that have a connection with $i$ (self included), $c_{ij}$ is a problem-specific normalization coefficient. It can be proven that Eq. \ref{eq:nodegcn} is equivalent to the original GCN Eq. \ref{eq:GCN} when $c_{ij}$ is the entry of the symmetrically normalized graph Laplacian $\tilde{D}^{-\frac{1}{2}} \tilde{A} \tilde{D}^{-\frac{1}{2}} $. Eq. \ref{eq:nodegcn} can be easily interpreted as that a node accepts messages from its neighbors \cite{gilmer2017neural}, by adding self-connection, a node is also considered a neighbor of itself.

Eq. \ref{eq:nodegcn} can be extended to multiple relations as Eq. \ref{eq:rgcn} \cite{schlichtkrull2018modeling}, where $r$ indicates a certain relation from a set of relations $R$ and ${N}_r(i)$ represents all the nodes that have relation $r$ with node $i$.
\begin{equation}\label{eq:rgcn}
h^{(k+1)}_i = \phi \left(\sum_{r\in R}\sum_{j\in {N}_r(i)} c_{ij}^r h_j^{(k)}W_r^{(k)} \right)
\end{equation}

The relational GCN formulated by Eq. \ref{eq:rgcn} is interpreted as that a node accepts messages from the nodes that have any relations with it. The message passing weights $W_r^{(k)}$ vary with different relations and different layers. In Eq. \ref{eq:rgcn}, note that there is a special relation in $R$ that deserves more attention, \ie the self-connection (denoted by $r_0$). We have $N_{r_0}(i)=\{i\}$, $c_{ii}^{r_0}=1$ if we consider each node equally accepts the self-contribution during information updating. Different from the original GCN Eqs. \ref{eq:GCN} and \ref{eq:nodegcn}, where all connections, including the self-connection, are considered equally, the relational GCN designs different message passing methods for different relations, including the self-connection.

We can also write Eq. \ref{eq:rgcn} in matrix form as Eq. \ref{eq:matrgcn}, where $A_r$ is a normalized adjacency matrix for relation $r$, for self-connection $r_0$, $A_{r_0}$ is an identity matrix. By Eq. \ref{eq:matrgcn}, the computation efficiency can be improved using sparse matrix multiplications.
\begin{equation}\label{eq:matrgcn}
H^{(k+1)} = \phi \left(\sum_{r\in R}A_r H^{(k)}W_r^{(k)}\right)
\end{equation}

Note that Eq. \ref{eq:rgcn} and \ref{eq:matrgcn} can be generalized to the situation of multi-relations between two nodes and the directed graphs. For multi-relations between two nodes, the message passing should be conducted multiple times, one for each relation. For directed graphs, the directed edges can be regarded as two relations, \ie the \textit{in} relation and the \textit{out} relation, thus there should be two different message passing methods corresponding to the message passing from the head node to tail node and from the tail to the head, respectively.

\begin{figure*}
  \centering
  \includegraphics[width=\linewidth,page=1]{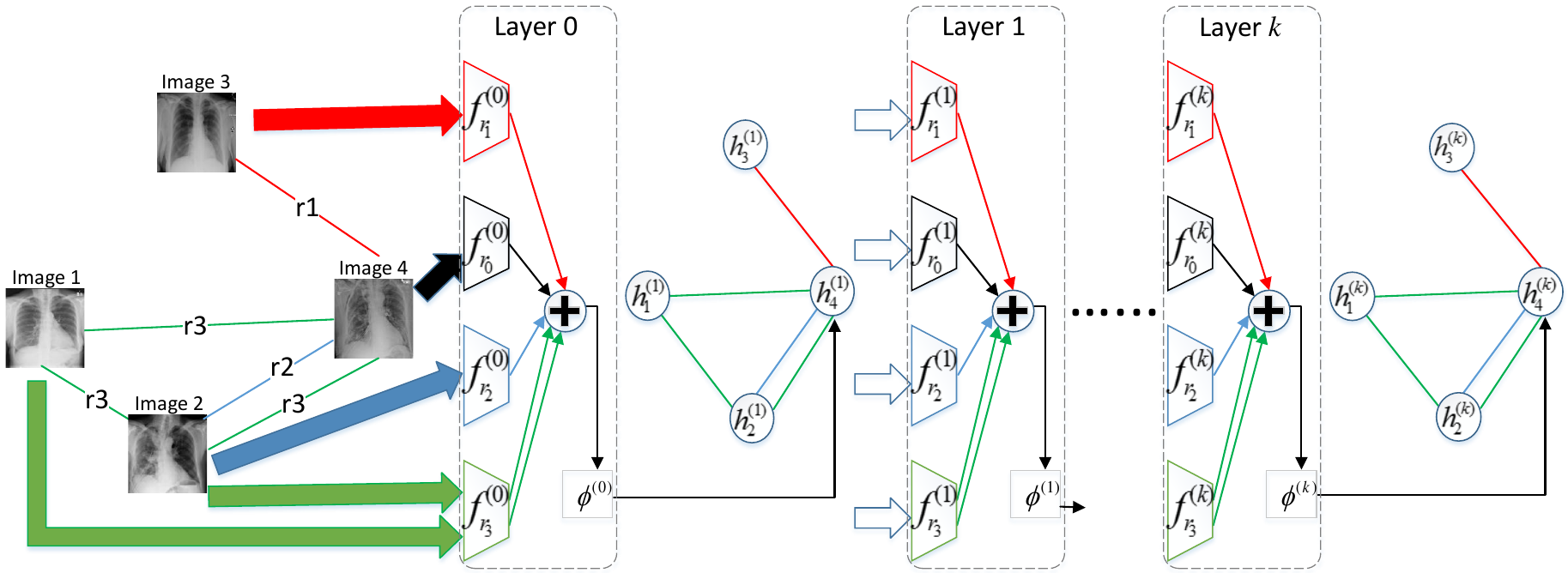}
  \caption{The propagation rule of ImageGCN. To avoid cumbersomeness, we only show the propagation of Image 4, the other images propagate in a similar rule. A dashed box is a GCN layer that consists of a number of message passing units (\ie $f_{r_i}^{(k)}$, corresponding to the number of relations) and a number of aggregators (\ie $\bigoplus$, corresponding to the number of images) followed by an activation function (\ie $\phi^{(k)}$). $r_1, r_2, r_3$ is the relations between images, $r_0$ is the self-connection relation. Colors indicate the propagation for different relations. $h_i^{(k)}$ indicates the representation of Image $i$ in layer $k$. In the propagation, the relations are preserved.}
  \label{fig:ImageGCN}
\end{figure*}

\subsection{ImageGCN}
Given the feature matrix of nodes(\ie $H^{(0)}$) and all the adjacency matrices for all relations between the nodes (\ie $A_r$), we can learn an informative representation for each node through Eq. \ref{eq:matrgcn}
However, Eq. \ref{eq:matrgcn} can not be directly extended to an ImageGraph like Fig. \ref{fig:ImageGraph}(a), where the original feature for each image is a 3-dimensional tensor ($channels \times width \times height $). If we flatten the tensor and use the linear transformation matrix $W_r^{(k)}$ for message passing, the transformation matrix will be extremely large, low efficiency and even low non-linear expressive capacity. To tackle this issue, in our ImageGCN, we propose to design flexible message passing methods between images as
\begin{equation}\label{eq:imagegcn}
H^{(k+1)} = \phi \left(\sum_{r\in R}A_r \cdot f_r^{(k)}(H^{(k)})\right)
\end{equation}
where $f_r^{(k)}$ is the kernel Message Passing Unit (MPU) corresponding to relation $r$ in layer $k$, $H^{(k)}$ is the input of the $k$th layer and the output of the $(k-1)$th layer, it can be a 4-dimensional tensor ($images \times channels \times width \times height $) that is the representations of the all images in the $k$th layer. $H^{(0)}$ is the original pixel-level input tensor of images. In the last layer, $H^{(k)}$ should be a matrix where each row corresponds to a distributed representation of an image.
The multiplication between a matrix ($A_r$) and a tensor ($f_r^{(k)}(H^{(k)})$) is involved in Eq. \ref{eq:imagegcn}. We expand the multiplication operation between matrices to a matrix $M$ and a tensor $T$ as
\begin{equation}\label{eq:tensormult}
[M \cdot T]_i = \sum_{j}M_{ij}T_j
\end{equation}
where $[M \cdot T]_i$ indicates result tensor's $i$th component, similarly $T_j$ is the $j$th component of $T$.

The propagation rule of ImageGCN can be illustrated in Fig. \ref{fig:ImageGCN}, where each node of the input ImageGraph gets a representation through a GCN layer, by stacking multiple GCN layers, each node could get an informative representation eventually.

\textbf{Inputs.}
An input ImageGraph consists of nodes (\ie images) and edges (\ie the relations between images). Each node has the inherent pixel-level features represented by a 3-dimensional tensor, initially. The edges corresponding to each relation are represented by an adjacency matrix which defines the neighbors of each node.

\textbf{ImageGCN Layer.} An ImageGCN layer contains a number of MPUs to do message passing between two connected nodes. An MPU corresponds to the message passing with respect to a type of relation. An ImageGCN layer also has an aggregator for each node to aggregate the received messages from its neighbors. An activation function (\eg ReLU) is applied to the aggregation to enhance the non-linear expressive capacity. In ImageGCN, MPUs can be designed as a multi-layer CNN architecture in the middle ImageGCN layers to filter informative messages, and linear MPUs can be used in the last layer to generate vectors to be aggregated for image representations.

\textbf{Propagation.} For each image (\eg Image 4 in Fig. \ref{fig:ImageGCN}), each of its neighbors are input to the corresponding MPU, the outputs are aggregated and then activated to generate the new representation of this image in the next layer. For each image, the propagation rule is

\begin{equation}\label{eq:nodeimagegcn}
h^{(k+1)}_i = \phi \left(\sum_{r\in R}\sum_{j\in {N}_r(i)} a_{ij}^r f_r^{(k)}(h_j^{(k)}) \right)
\end{equation}
where $a_{ij}^r$ is the entry of normalized adjacency matrix of relation $r$. Eq. \ref{eq:nodeimagegcn} is equivalent to Eq. \ref{eq:imagegcn} and can be seen as a generalization of Eq. \ref{eq:rgcn}.

Fig. \ref{fig:ImageGCN} illustrates the propagation rule in Eq \ref{eq:nodeimagegcn}. It shows how the image representations are generated by ImageGCN. Taking Image 4 as an example, the whole propagation process is as follows.
\begin{enumerate}
    \item 	There are 4 types of relations in the ImageGraph including the self-connection relation, denoting as $r_0$, $r_1$, $r_2$, $r_3$, thus, 4 types of MPUs are designed in each layer of ImageGCN (\eg, $f_{r_0}^{(k)}$, $f_{r_1}^{(k)}$,  $f_{r_2}^{(k)}$, $f_{r_3}^{(k)}$ in layer $k$) corresponding to the 4 types of relations.
	\item   Each image that has a relation with Image 4 is fed to the corresponding MPU to get a temporary representation. With regard to Image 4, Image 3 has relation $r_1$; Image 2 has relation $r_2$ and $r_3$; Image 1 has relation $r_3$, thus, Image 3 is fed to  $f_{r_1}^{(0)}$; Image 2 is fed to  $f_{r_2}^{(0)}$ and $f_{r_3}^{(0)}$; Image 1 is fed to $f_{r_3}^{(0)}$; Image 4 itself is fed to self-connection MPU $f_{r_0}^{(0)}$.
	\item   All the output temporary representations are aggregated and then input to an activation function to get a single representation which is regarded as the representation of Image 4 in Layer 0.
	\item   The graph with all image representations in Layer 0 as the features are input to Layer 1 with similar propagation rule. Repeat steps 2) and 3) in Layer 2 and get the representation in Layer 2.
	\item   Recursively, the representation of images in the last layer are recorded for the downstream tasks.
\end{enumerate}

\begin{figure}[!t]
  \centering
  \subfloat[Partial parameter sharing]{ \includegraphics[width=.45\linewidth, height=.45\linewidth,page=1]{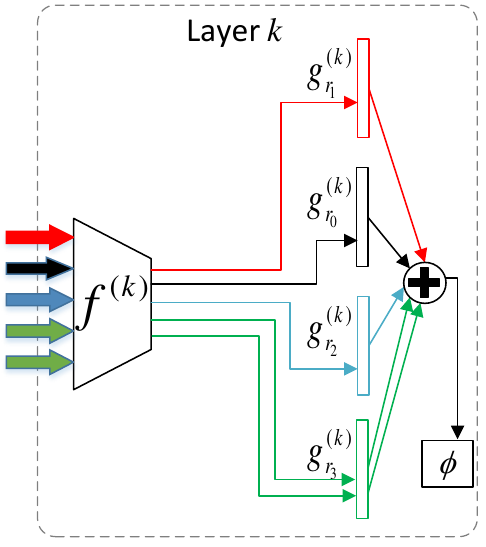} \label{fig:partialps}}
  \hspace{0.1in}
  \subfloat[All parameter sharing]{ \includegraphics[width=.45\linewidth,height=.45\linewidth,page=2]{latex/PPS.pdf} \label{fig:totalps}}
  \caption{The parameter sharing schemes. (a) Partial parameter sharing; (b) All parameter sharing.  }
  \label{fig:parametersharing}
\end{figure}

\textbf{Partial Parameter Sharing.} \label{sec:parametersharing}
Because each relation has an MPU, an issue with applying Eq. \ref{eq:imagegcn} to a ImageGraph with many relation types is that the number of parameters would grow rapidly with the number of relations. This will lead to a very large model that is not easy to train with limited computing and storage resources, especially for MPUs with multi-layer neural networks.

To address this issue, we introduce the partial parameter sharing (PPS) scheme between MPUs. With PPS, The MPUs share most of the parameters to reduce the total number of parameters. In our design, the same CNN architecture is applied to all MPUs in the same layer, all the parameters are shared between these MPUs except for the last parameter layer where the parameters are used to make the message passing rule different for different relations, see Fig. \ref{fig:partialps} for an ImageGCN layer with PPS. Thus, the message passing rule Eq. \ref{eq:imagegcn} can be further refined as:

\begin{equation}\label{eq:imageppsgcn}
H^{(k+1)} = \phi \left(\sum_{r\in R}A_r \cdot g_r^{(k)}(f^{(k)}(H^{(k)}))\right)
\end{equation}
where $f^{(k)}$ is shared by all relations, only $g_r^{(k)}$ that has only a few parameters determines the different message passing methods for different relations. Also, we can further share all the parameters between all MPUs, that is, assigning the same message passing rule to different relations, \ie all parameter sharing (APS) in Fig. \ref{fig:totalps}. However, APS will reduce the multiple relations to a single relation, thus reduce the model's expressive capacity, our experimental results in also demonstrate the less effectiveness of APS than PPS.

\subsection{Training Strategies}
\subsubsection{Loss function.}
The loss function relies on the downstream task. Specifically, for a classic node classification task, we can use a softmax activation function in the last layer and minimize the cross-entropy loss on all labeled nodes. For multi-label classifications, the loss function can be design as in our experiments in Section \ref{sec:loss}.

\subsubsection{Batch propagation.}\label{sec:bp}
Equation \ref{eq:imageppsgcn} requires all nodes in the graph being present during training, \ie it can not support propagation in batch. This is difficult to scale out to a large graph with high-dimensional node features, which is common in computer vision. One may want to simply construct a subgraph in a batch, this usually causes no edges in a batch if the graph is sparse. GraphSAGE \cite{hamilton2017inductive} was designed to address this issue for single relational graphs. Inspired by GraphSAGE, we introduce an inductive batch propagation algorithm for multi-relational ImageGraphs in Algorithm \ref{al:batchtrain}. For each sample $v$ in a batch, for each relation $r$, we randomly sample $n$ neighbors of $v$ to pass message to $v$ with relation $r$ in a layer (Line \ref{al:sample}). The union of the sampled neighbors and the samples in the batch are considered as a new batch for the next layer (Line \ref{al:batchinit} to \ref{al:batchneighbor}). For a $k$-layer ImageGCN, the neighbor sampling should be repeated $k$ times to reach the $k$th order neighbors of the initial batch (Line \ref{al:begink} to \ref{al:endk}). We construct the subgraph based on the final batch (Line \ref{al:loadneighbor} to \ref{al:subgraph}, $B^{(0)}$ is the final batch). In each ImageGCN layer, the message passing is conducted inside the subgraph (Line \ref{al:messpass}). Note that the image features can be in persistent storage, and are loaded when a batch and the neighbors of images in the batch are sampled (Line \ref{al:loadneighbor}), This is important to reduce memory requirement for large-scale graphs or graphs with high-dimensional features in the nodes.

\begin{algorithm}[t]
\caption{ ImageGCN batch propagation algorithm.  }
\label{al:batchtrain}
\begin{algorithmic}[1]
\REQUIRE ~~ \\
graph node set $V$ and the mini-batch $B$;  \\
relation adjacency matrix $A_r, r\in R$;  \\
input image features $X$ (can be stored externally);    \\
network depth $K$;   \\
number of neighbors to sample for each node and each relation $n$.  \\
\ENSURE The representation of all samples in $B$
\STATE $B^{(K)} \leftarrow B$
\FOR  {$k=K \cdots 1$ }   \label{al:begink}
\STATE $B^{(k-1)}\leftarrow B^{(k)}$  \label{al:batchinit}
    \FOR {$r \in R$}
        \FOR {$v \in B^{(k)}$}
        \STATE $N_r(v) \leftarrow$ the neighbor set of $v$  based on $A_r$
        \STATE $n\leftarrow \min(n, |N_r(v)|)$
        \STATE $S\leftarrow$ random $n$ samples from $N_r(v)$  without replacement  \label{al:sample}
        \STATE $B^{(k-1)} \leftarrow B^{(k-1)}\cup S$
        \ENDFOR
    \ENDFOR   \label{al:batchneighbor}
\ENDFOR   \label{al:endk}
\STATE load the features $H^{(0)}$ for $B^{(0)}$ from $X$  \label{al:loadneighbor}
\FOR {$r\in R$}
\STATE $A_r^B\leftarrow$ the sub-matrix corresponding to $B^{(0)}$ in the adjacency matrix $A_r$
\ENDFOR  \label{al:subgraph}
\STATE execute Eq. \ref{eq:imageppsgcn} with $A_r=A_r^B$, for $k=0,\cdots,K-1$  \label{al:messpass}
\STATE extract the representations of samples in $B$ from $H^{(K)}$
\end{algorithmic}
\end{algorithm}

In test process, for a test image, if the training samples are all no longer available or the relations between test image  and the training samples are unknown, the only available relation is self-connection, thus the propagation rule in Eq. \ref{eq:nodeimagegcn} can be simplified as
\begin{equation}\label{eq:spimagegcn}
h^{(k+1)}_i = \phi \left( f_{r_0}^{(k)}(h_i^{(k)}) \right)
\end{equation}
where $f_{r_0}^{(k)}(\cdot)$ is the MPU corresponding to self-connection in the $k$th layer which is trained in the training process. If the test image has any relations with any available training images, the test image and the relations are added to the adjacency matrices $A_r$. The batch propagation algorithm \ref{al:batchtrain} can be directly applied for test image representation.

\section{Experiments}
\subsection{Data sets}
We test ImageGCN for disease identification on ChestX-ray14 dataset \cite{wang2017chestxray}, CheXpert dataset \cite{irvin2019chexpert} and MIMIC-CXR dataset \cite{johnson2019mimic}. Since ChestX-ray14 dataset also provides the a subset of labelled pathology bounding box annotations, we perform disease localization on ChestX-ray14 dataset.
\subsubsection{ChestX-ray14 Dataset}
ChestX-ray14 dataset consists of 112,120 frontal-view CXR images of 30,805 patients related with 14 thoracic disease labels. The labels are mined from the associated radiological reports using natural language processing, and are expected to have accuracy$>$90\%. Out of the 112,120 CXR images, 51,708 contains one or more pathologies. The remaining 60,412 images are considered normal. ChestX-ray14 dataset also provides the patients information (\eg age, gender) and the image information (\eg view) for a CXR image based on which we construct the ImageGraph.

\subsubsection{CheXpert Dataset}
CheXpert dataset consists of 224,316 chest radiographs of 65,240 patients labeled for the presence of 13 common chest radiographic observations as positive, negative, uncertain. The labels are also extracted from the associated radiological reports using natural language processing tools. CheXpert dataset also provides the patients information (\eg age, gender) and the image information (\eg view) for a CXR image based on which we construct the ImageGraph.

\subsubsection{MIMIC-CXR Dataset}
MIMIC-CXR dataset consists of 377,110 CXR images of 65,379 patients corresponding to 227,835 radiographic studies labeled for the presence of 13 common chest radiographic observations as positive, negative, uncertain. The labels are also extracted from the associated radiological reports using natural language processing tools. MIMIC-CXR dataset does not provide the patients age and gender, thus we construct the ImageGraph based on the person information and view information.

For all the datasets, we randomly split the dataset into training, validation and test set by the ratio 7:1:2. We regard the provided labels as ground truth to train the model on training set and evaluate it on test set. We do not apply any data augmentation techniques. Each image in the dataset is resized to $256\times 256$, and then cropped to $224 \times 224$ at the center for fast processing. We normalized the image by mean ([0.485; 0.456; 0.406]) and standard deviation ([0.229; 0.224; 0.225]) of the images from ImageNet \cite{deng2009imagenet}.

\subsection{Graph Construction} \label{sec:graphdefine}
With the advance of precision medicine, individualization, age and gender are increasingly recognized as patient disease and outcome associating factors \cite{regitz2012sex,kohane2015ten}. For chest X-ray, different views offer different diagnostic information regarding the same parts of the body \cite{ittyachen2017forgotten}. Thus, besides the self-connection, we take the 4 types of relations into account to construct an ImageGraph. (1) \textbf{Person relation,} if two images come from the same person, a person relation exists. (2) \textbf{Age relation,} if the two images come from the persons of the same age when the CXR were taken, an age relation exists. (3) \textbf{Gender relation,} if the owners of two images have the same gender, a gender relation exists. (4) \textbf{View relation,} if two CXR images were taken with the same view position (PosteroAnterior or AnteroPosterior ), a view relation exists. For MIMIC-CXR dataset, since the patients' age and gender information is missing, we only consider the person relation and view relation.

The four relations are all reflexive, symmetric and transitive, thus each relation corresponds to a cluster graph that consists of a number of disjoint complete subgraphs. Person relation usually implies gender relation but can not imply age relation, because a person can take several CXR images at different ages. The adjacency matrix of each relation is a diagonal block matrix. Our ImageGCN is built on this multi-relational graph. The adjacency matrices are normalized in advance. Note that because the self-connection relation is considered separately, The adjacency matrices do not need to add self-connection.

\begin{table*}[!t]
  \scriptsize
  \centering
    \caption{The AUROC results of various models to classify the 14 diseases on ChestX-ray14 dataset. The upper part is the results from our experiments using different base models and the corresponding ImageGCN models. The middle part is the state-of-the-art results reported in the recent studies. The lower part is the significance test results of repeated measures ANOVA and the post-hoc tests. *, ** and *** denote the significance levels of 0.1, 0.05 and 0.01, respectively. For each disease, the best results are bolded. The red text means our ImageGCN can perform better than the corresponding two baseline models.
    }
    \begin{tabular}{l|cccccccccccccc|c|l}
    \toprule
          & Atel & Card & {Effu} & {Infi} & {Mass} & {Nodu} & {Pneu1} & {Pneu2} & {Cons} & {Edem} & {Emph} & {Fibr} & PT & {Hern} & {mean} & MR \\
    \midrule
    A-GCN-PPS (ours) & 0.781 & \textcolor{red}{0.899} & \textcolor{red}{0.865} & 0.701 & \textcolor{red}{0.813} & \textcolor{red}{0.721} & \textcolor{red}{0.718} & \textcolor{red}{0.881} & 0.788 & 0.888 & \textcolor{red}{0.882} & \textcolor{red}{0.804} & \textcolor{red}{0.778} & 0.904 & \textcolor{red}{0.816} & \textcolor{red}{6.73} \\
    A-GCN-APS & 0.739 & 0.876 & 0.815 & 0.671 & 0.799 & 0.704 & 0.679 & 0.857 & 0.762 & 0.846 & 0.863 & 0.792 & 0.765 & 0.910 & 0.791 & 13.73 \\
    AlexNet & 0.782 & 0.895 & 0.863 & 0.705 & 0.781 & 0.714 & 0.716 & 0.869 & 0.790 & \textbf{0.889} & 0.876 & 0.799 & 0.773 & 0.899 & 0.811 & 7.93 \\
    \midrule
    R-GCN-PPS (ours) & 0.785 & \textcolor{red}{0.890} & \textcolor{red}{0.868} & \textcolor{red}{0.699} & \textcolor{red}{0.824} & \textcolor{red}{0.739} & \textcolor{red}{0.723} & \textcolor{red}{0.895} & 0.790 & 0.887 & \textcolor{red}{0.911} & \textcolor{red}{0.819} & \textcolor{red}{0.786} & \textcolor{red}{0.941} & \textcolor{red}{0.826} & \textcolor{red}{4.47}\\
    R-GCN-APS & 0.741 & 0.861 & 0.822 & 0.680 & 0.819 & 0.728 & 0.684 & 0.873 & 0.768 & 0.852 & 0.889 & 0.790 & 0.751 & 0.908 & 0.798 & 12.27\\
    ResNet50 & 0.789 & 0.889 & 0.863 & 0.698 & 0.807 & 0.723 & 0.714 & 0.876 & 0.791 & 0.888 & 0.899 & 0.799 & 0.772 & 0.933 & 0.817 & 6.93 \\
    \midrule
    V-GCN-PPS (ours) & \textcolor{red}{\textbf{0.802}} & \textcolor{red}{0.894} & \textcolor{red}{\textbf{0.874}} & \textcolor{red}{0.702} & \textcolor{red}{\textbf{0.843}} & \textcolor{red}{0.768} & {0.715} & \textcolor{red}{\textbf{0.900}} & \textcolor{red}{0.796} & \textcolor{red}{0.883} & \textcolor{red}{0.915} & \textcolor{red}{0.825} & \textcolor{red}{\textbf{0.791}} & \textcolor{red}{\textbf{0.943}} & \textcolor{red}{\textbf{0.832}} & \textcolor{red}{\textbf{2.60}}\\

    V-GCN-APS & 0.754 & 0.871 & 0.826 & 0.676 & 0.820 & 0.737 & 0.688 & 0.872 & 0.769 & {0.839} & 0.894 & 0.789 & {0.770} & 0.926 & 0.802 & 11.07\\
    VGGNet16BN & 0.796 & 0.893 & 0.872 & 0.700 & 0.831 & 0.756 & 0.717 & 0.882 & 0.794 & 0.878 & 0.909 & 0.799 & 0.785 & 0.923 & 0.824 & 4.60 \\
    \midrule\midrule
    Wang et al. \cite{wang2017chestxray}  & 0.716 & 0.807 & 0.784 & 0.609 & 0.706 & 0.671 & 0.633 & 0.806 & 0.708 & 0.835 & 0.815 & 0.769 & 0.708 & 0.767 & 0.738 & 17.33 \\
    Yao et al. \cite{yao2017learning}  & 0.772 & 0.904 & 0.859 & 0.695 & 0.792 & 0.717 & 0.713 & 0.841 & 0.788 & 0.882 & 0.829 & 0.767 & 0.765 & 0.914 & 0.803 & 10.80 \\
   Li et al. \cite{li2018thoracic}  & 0.800 & 0.870 & 0.870 & 0.700 & 0.830 & 0.750 & 0.670 & 0.870 & \textbf{0.800} & 0.880 & 0.910 & 0.780 & 0.760 & 0.770 & 0.804 & 8.40 \\
    Kumar et al. \cite{kumar2018boosted} & 0.762 & \textbf{0.913} & 0.864 & 0.692 & 0.750 & 0.666 & 0.715 & 0.859 & 0.784 & 0.888 & 0.898 & 0.756 & 0.774 & 0.802 & 0.794 & 10.67\\
    Tang et al. \cite{tang2018attention} & 0.756 & 0.887 & 0.819 & 0.689 & 0.814 & 0.755 & 0.729 & 0.85  & 0.728 & 0.848 & 0.906 & 0.818 & 0.765 & 0.875 & 0.803 & 10.33\\
    Shen et al. \cite{shen2018dynamic} & 0.766 & 0.801 & 0.797 & \textbf{0.751} & 0.76  & 0.741 & \textbf{0.778} & 0.800 & 0.787 & 0.82  & 0.773 & 0.765 & 0.759 & 0.748 & 0.775 & 13.40 \\
    Mao et al. \cite{mao2018deep}   & 0.750 & 0.869 & 0.810 & 0.687 & 0.782 & 0.726 & 0.695 & 0.845 & 0.728 & 0.834 & 0.870 & 0.798 & 0.758 & 0.877 & 0.788 & 14.33\\
    Guan et al. \cite{guan2018multi}  & 0.781 & 0.883 & 0.831 & 0.697 & 0.83  & 0.764 & 0.725 & 0.866 & 0.758 & 0.853 & {0.911} & {0.826} & 0.78  & 0.918 & 0.816 & 7.00\\
    Liu et al. \cite{liu2020semi} & 0.773 & 0.889 & 0.821 & 0.710 & 0.829 & \textbf{0.770} & 0.713 & 0.869 & 0.749 & 0.847 & \textbf{0.934} & \textbf{0.845} & 0.773 & 0.925 & 0.818 & 6.93  \\
    \midrule\midrule

{p-val (RM-ANOVA)}      & ***         & ***          & ***      & ***          & *    & *      & ***       & ***          & ***           & ***   & ***       & **       & **                  &        & ***  \\
{PPS $>$ APS } & ***         & ***          & ***      & ***          & *    & **     & ***       & ***          & ***           & ***   & ***       & **       & **                  &        & ***  \\
{PPS $>$  base} &             & *            & **       &              & **   & **     &           & ***          &               &       & **        & *        & **                  & *      & **  \\
\bottomrule

    \end{tabular}   %

    Atel: Atelectasis; Card: Cardiomegaly; Effu: Effusion; Infi: Infiltration; Nodu: Nodule;	Pneu1: Pneumonia;	Pneu2:Pneumothorax;	Cons: Consolidation;	Edem: Edema;	Emph: Emphysema;	Fibr: Fibrosis;	PT:Pleural Thickening;	Hern: Hernia. MR=mean rank

  \label{tab:AUROC}%
\end{table*}%

\subsection{MPU design}
Since the ImageGraph in our experiments is a cluster graph for each relation, each node can reach other reachable nodes by 1 step, one-layer ImageGCN is enough to catch the structure information of an image node. Stacking multiple GCN layers would result in over-smoothing issues \cite{li2018deeper}. For the one-layer ImageGCN, we design the MPUs in our experiments as a deep CNN architecture to catch high-level visual information. According to PPS in Fig. \ref{fig:parametersharing}, each MPU consists of two parts: the sharing part $f^{(0)}$ and the private part $g_r^{(0)}$.

\subsubsection{The sharing part} The sharing part of the MPUs consists of the feature layers of a pre-trained CNN architecture, a transition layer and a global pooling layer, sequentially. For a pre-trained model, we discard the high-level fully-connected layers and classification layers and only keep the remaining feature layers as the first component of the sharing part.
We append a transition layer and then a global pooling layer to the end of the feature layers to construct the sharing part.
The transition layer consists of a convolutional layer, a batch normalization layer \cite{ioffe2015batch} and a ReLu layer sequentially. In the transition layer, we let the convolutional layer have 1024 filters with kernel size $3\times 3$ to transform the output of previous layers into a uniform number (\ie 1024 in our experiment) of feature maps which is used to generate the heatmap for disease localization. The global pooling layer pools the generated 1024 feature maps to a 1024-dimensional vector with a kernel size equal to the feature map's size. Thus, by the sharing part of MPUs, an image is transformed to a 1024-dimensional vector. We test the feature layers of three different pre-trained CNN architectures independently in our experiments, \ie AlexNet \cite{krizhevsky2012imagenet}, VGGNet16 with batch normalization (VGGNet16BN) \cite{simonyan2014very}, and ResNet50 \cite{he2016deep}. All the CNN architectures are pre-trained on ImageNet and fine-tuned with our classification task on ChestX-ray14 dataset.

\subsubsection{The private part} The private part accepts the output of the sharing part and outputs an embedding to the aggregator. For each relation, we use a linear layer (with different parameters) as the private part to transform the 1024-dimensional vector from the sharing part to a $d$-dimensional vector, {$d$ is the number of classes, $d=14$ for ChestX-ray14 dataset and $d=13$ for CheXpert and MIMIC-CXR datasets }. For an image, the d-dimensional vectors from its neighbors are aggregated and fed to a sigmoid activation function to generate its probabilities corresponding to the d classes. With the method of class activation mapping (CAM)  \cite{zhou2016learning}, the weights of the private linear layer of self-connection combined with the activations of the transition layer in the sharing part can generate a heatmap for the disease location task.

All the learnable parameters of the ImageGCN model are contained in these two parts, the sharing part corresponds to the feature layers of a pre-trained architecture, and the private part contains 5 linear layers corresponding to the 4 relations and self-connection. Though only a part of the pre-trained model, \eg AlexNet, is incorporated in an MPU, we call it an AlexNet MPU for convenience, similarly, VGGNet16BN MPU and ResNet50 MPU. For each MPU type (\eg AlexNet), we use two baselines to evaluate our model, ImageGCN with all parameter sharing (APS) and the basic pre-trained model (AlexNet) fine-tuned in the dataset. In the following statement in this paper, we use A-GCN-PPS to denote the ImageGCN with AlexNet MPUs and partial parameter sharing, similarly V-GCN-PPS for VGGNet16BN MPUs and R-GCN-PPS for ResNet50 MPUs.

\subsection{Experimental settings}

\subsubsection{Loss function.} \label{sec:loss} For multi-label classification on ChestX-ray14, the true label of each CXR image is a 14-dimensional binary vector $y = [y_1, \cdots, y_{14}]; y_i\in \{0,1\} $ where $y_i=1$ denotes the corresponding disease is present and $y_i=0$ for absence. An all zero vector represents ``No Findings'' in the 14 diseases. Due to the high sparsity of the label matrix, we use the weighted cross entropy loss as Wang et al. \cite{wang2017chestxray} did, where each sample with true labels $y$ and output probabilities $p$ has the loss

\begin{equation}\label{eq:Mloss}
l(p,y) = -\sum_{y_i=1}  \frac{N_n}{N_p}\log{p_{i}} -\sum_{y_i=0}\log{(1-p_{i})}
\end{equation}
where $N_n$ and $N_p$ are the number of `0's and `1's in a mini-batch respectively. The loss of images in a mini-batch are averaged as the loss of the batch.

\subsubsection{Hyperparameters.} We set the batch size to 16. 1 neighbor is sampled for each image and each relation. All the models are trained using Adam optimizer \cite{kingma2014adam} with parameters $lr=10^{-5},\beta=(0.9, 0.999), eps=10^{-8}, weight\_decay=0.01$. We terminate the training procedure when it reaches 10 epochs. In each epoch, the model with the best classification performance on the validation set is saved for evaluation.

\begin{table*}[htbp]
\scriptsize
  \centering
    \caption{{The AUPRC results of various models to classify the 14 diseases on ChestX-ray14 dataset. The upper part is the results from our experiments using different base models and the corresponding ImageGCN models. The lower part is the significance test results of repeated measures ANOVA and the post-hoc tests. *, ** and *** denote the significance levels of 0.1, 0.05 and 0.01, respectively. For each disease, the best results are bolded. The red text means our ImageGCN can perform better than the corresponding two baseline models. }
    }
\begin{tabular}{l|cccccccccccccc|c}
\toprule
          & Atel & Card & {Effu} & {Infi} & {Mass} & {Nodu} & {Pneu1} & {Pneu2} & {Cons} & {Edem} & {Emph} & {Fibr} & PT & {Hern} & {mean}  \\
    \midrule
A-GCN-PPS & 0.275 & {\color{red} 0.256} & {\color{red} 0.479} & {\color{red} \textbf{0.347}} & {\color{red} 0.237} & {\color{red} 0.172} & {\color{red} 0.037} & {\color{red} 0.303} & {\color{red} 0.132} & {\color{red} 0.145} & {\color{red} 0.241} & {\color{red} 0.102} & {\color{red} 0.120} & 0.059 & {\color{red} 0.208} \\
A-GCN-APS & 0.243 & 0.212 & 0.384 & 0.301 & 0.224 & 0.165 & 0.028 & 0.284 & 0.123 & 0.116 & 0.221 & 0.071 & 0.107 & 0.070 & 0.182 \\
AlexNet & 0.280 & 0.251 & 0.468 & 0.335 & 0.229 & 0.163 & 0.032 & 0.300 & 0.122 & 0.139 & 0.181 & 0.088 & {0.110} & 0.055 & 0.197 \\
\midrule
R-GCN-PPS & 0.281 & {\color{red} 0.250} & {\color{red} 0.478} & {\color{red} 0.338} & {\color{red} 0.238} & {\color{red} 0.186} & {\color{red} \textbf{0.038}} & {\color{red} 0.344} & {\color{red} \textbf{0.142}} & {\color{red} 0.146} & {\color{red} 0.354} & {\color{red} 0.094} & {\color{red} 0.117} & 0.094 & {\color{red} 0.222} \\
R-GCN-APS & 0.228 & 0.197 & 0.378 & 0.309 & 0.198 & 0.175 & 0.025 & 0.277 & 0.115 & 0.109 & 0.250 & 0.076 & 0.105 & 0.144 & 0.185 \\
ResNet50 & 0.288 & 0.243 & 0.476 & 0.330 & 0.237 & 0.163 & 0.033 & 0.311 & 0.127 & 0.144 & 0.332 & 0.091 & 0.117 & 0.077 & 0.212 \\
\midrule
V-GCN-PPS & {\color{red} \textbf{0.313}} & {\color{red} \textbf{0.267}} & {\color{red} \textbf{0.511}} & {\color{red} 0.340} & {\color{red} \textbf{0.328}} & {\color{red} \textbf{0.225}} & {\color{red} 0.037} & {\color{red} \textbf{0.351}} & {\color{red} 0.141} & {\color{red} \textbf{0.154}} & {\color{red} \textbf{0.357}} & {\color{red} \textbf{0.105}} & {\color{red} \textbf{0.139}} & {\color{red} \textbf{0.304}} & {\color{red} \textbf{0.255}} \\
V-GCN-APS & 0.272 & 0.216 & 0.414 & 0.317 & 0.291 & 0.212 & 0.035 & 0.313 & 0.124 & 0.137 & 0.258 & 0.103 & 0.131 & 0.263 & 0.220 \\
VGGNet16BN & 0.297 & 0.248 & 0.497 & 0.332 & 0.262 & 0.206 & 0.037 & 0.328 & 0.135 & 0.151 & 0.334 & 0.096 & 0.125 & 0.079 & 0.223 \\
\midrule\midrule
p-val (RM-ANOVA) & ** & *** & *** & *** &  & ** & * & ** & ** & *** &  &  & * &  & ** \\
PPS \textgreater APS & *** & *** & *** & ** & ** & ** & * & * & ** & ** & * & * & ** &  & *** \\
PPS \textgreater base &  & * & * & ** &  & ** &  & * & ** & ** & * & * & * &  & * \\
\bottomrule
\end{tabular}
 \label{tab:AUPRC}
\end{table*}

\begin{table*}[htbp]
\scriptsize
  \centering
    \caption{{The AUROC and AUPRC results of various models to classify the 13 classes (12 diseases and Support Devices) on CheXpert dataset. The upper part is the AUROC results and the lower part is the AUPRC results. *, ** and *** denote the significance levels of 0.1, 0.05 and 0.01, respectively. For each disease, the best results are bolded. The red text means our ImageGCN can perform better than the corresponding two baseline models. }
    }
\begin{tabular}{c|l|ccccccccccccc|c}
\toprule
\multicolumn{1}{l}{} &  & EC & Card & AO & LL & Edem & Cons & Pneu1 & Atel & Pneu2 & PE & PO & Frac & SD & mean \\
\midrule
 & A-GCN-PPS & {\color{red} 0.675} & {\color{red} 0.860} & {\color{red} 0.716} & {\color{red} 0.750} & {\color{red} 0.850} & {\color{red} 0.721} & {\color{red} 0.732} & {\color{red} 0.700} & 0.833 & {\color{red} 0.867} & {\color{red} 0.783} & 0.743 & {\color{red} 0.839} & {\color{red} 0.775} \\
 & A-GCN-APS & 0.665 & 0.843 & 0.692 & 0.769 & 0.818 & 0.689 & 0.722 & 0.687 & 0.834 & 0.822 & 0.779 & 0.772 & 0.809 & 0.762 \\
 & AlexNet & 0.661 & 0.855 & 0.713 & 0.740 & 0.847 & 0.719 & 0.725 & 0.700 & 0.811 & 0.865 & 0.771 & 0.724 & 0.832 & 0.766 \\
\cmidrule{2-16}
 & R-GCN-PPS & {\color{red} 0.678} & {\color{red} \textbf{0.863}} & {\color{red} \textbf{0.724}} & {\color{red} 0.779} & {\color{red} \textbf{0.858}} & {\color{red} \textbf{0.735}} & {\color{red} \textbf{0.766}} & {\color{red} \textbf{0.712}} & {\color{red} 0.861} & {\color{red} \textbf{0.883}} & {\color{red} 0.801} & 0.785 & {\color{red} \textbf{0.859}} & {\color{red} \textbf{0.793}} \\
 & R-GCN-APS & 0.671 & 0.845 & 0.700 & 0.777 & 0.825 & 0.715 & 0.742 & 0.695 & 0.856 & 0.831 & 0.794 & 0.787 & 0.818 & 0.774 \\
 & ResNet50 & 0.662 & 0.856 & 0.721 & 0.743 & 0.854 & 0.729 & 0.738 & 0.708 & 0.825 & 0.877 & 0.775 & 0.730 & 0.849 & 0.774 \\
\cmidrule{2-16}
 & V-GCN-PPS & {\color{red} \textbf{0.697}} & {\color{red} \textbf{0.863}} & {\color{red} 0.721} & {\color{red} \textbf{0.796}} & {\color{red} 0.857} & {\color{red} 0.728} & {\color{red} 0.760} & {\color{red} 0.710} & 0.861 & {\color{red} 0.881} & 0.795 & 0.774 & {\color{red} 0.856} & {\color{red} 0.792} \\
 & V-GCN-APS & 0.667 & 0.844 & 0.706 & 0.770 & 0.826 & 0.717 & 0.757 & 0.701 & \textbf{0.862} & 0.831 & \textbf{0.812} & \textbf{0.812} & 0.819 & 0.779 \\
 & VGGNet16BN & 0.658 & 0.852 & 0.719 & 0.739 & 0.850 & 0.719 & 0.739 & 0.704 & 0.823 & 0.872 & 0.766 & 0.719 & 0.845 & 0.770 \\
\cmidrule{2-16}\cmidrule{2-16}
 & p-val (ANOVA) & ** & *** & *** & * & *** & * & * & ** & *** & *** & * & ** & *** & ** \\
 & PPS $>$ APS & * & *** & *** &  & *** & ** & * & ** &  & *** &  &  & *** & *** \\
\multirow{-14}{*}{AUROC} & PPS $>$ base & ** & ** & ** & * & ** & * & ** & * & ** & * & ** & ** & *** & ** \\
\bottomrule\toprule

 & A-GCN-PPS & {\color{red} 0.119} & {\color{red} 0.528} & {\color{red} 0.668} & {\color{red} 0.179} & {\color{red} 0.628} & {\color{red} 0.168} & 0.089 & {\color{red} 0.308} & 0.404 & {\color{red} 0.807} & 0.079 & 0.134 & {\color{red} 0.835} & {\color{red} 0.380} \\
 & A-GCN-APS & 0.111 & 0.483 & 0.645 & 0.126 & 0.576 & 0.160 & 0.107 & 0.296 & 0.410 & 0.747 & 0.095 & 0.173 & 0.806 & 0.364 \\
 & AlexNet & 0.117 & 0.524 & 0.661 & 0.118 & 0.628 & 0.168 & 0.094 & 0.307 & 0.359 & 0.805 & 0.068 & 0.118 & 0.825 & 0.369 \\
\cmidrule{2-16}
 & R-GCN-PPS & {\color{red} 0.123} & {\color{red} \textbf{0.543}} & {\color{red} \textbf{0.673}} & {\color{red} 0.213} & {\color{red} \textbf{0.648}} & {\color{red} \textbf{0.186}} & 0.121 & {\color{red} 0.319} & {\color{red} \textbf{0.471}} & {\color{red} \textbf{0.825}} & 0.101 & 0.185 & {\color{red} \textbf{0.849}} & {\color{red} 0.404} \\
 & R-GCN-APS & 0.115 & 0.489 & 0.654 & 0.175 & 0.590 & 0.182 & 0.132 & 0.308 & 0.453 & 0.756 & 0.131 & 0.213 & 0.813 & 0.385 \\
 & ResNet50 & 0.121 & 0.528 & 0.671 & 0.125 & 0.645 & 0.177 & 0.096 & 0.314 & 0.395 & 0.822 & 0.072 & 0.123 & 0.842 & 0.379 \\
\cmidrule{2-16}
 & V-GCN-PPS & {\color{red} \textbf{0.137}} & {\color{red} 0.534} & {\color{red} 0.671} & {\color{red} \textbf{0.258}} & {\color{red} 0.644} & {\color{red} 0.183} & 0.116 & {\color{red} \textbf{0.320}} & 0.460 & {\color{red} 0.824} & 0.095 & 0.183 & {\color{red} \textbf{0.849}} & {\color{red} \textbf{0.406}} \\
 & V-GCN-APS & 0.114 & 0.495 & 0.661 & 0.154 & 0.598 & 0.174 & \textbf{0.158} & 0.319 & 0.461 & 0.754 & \textbf{0.141} & \textbf{0.244} & 0.815 & 0.391 \\
 & VGGNet16BN & 0.108 & 0.516 & 0.668 & 0.119 & 0.633 & 0.165 & 0.098 & 0.308 & 0.382 & 0.812 & 0.072 & 0.115 & 0.839 & 0.372 \\
\cmidrule{2-16}\cmidrule{2-16}

 & p-val (RM-ANOVA) &  & *** & *** & ** & *** &  & * &  & *** & *** & ** & ** & *** & ** \\
 & PPS $>$ APS & * & *** & ** & ** & *** & ** &  & * &  & *** &  &  & *** & *** \\
\multirow{-14}{*}{AUPRC} & PPS $>$  base &  & * & ** & ** &  &  &  & * & ** &  & ** & ** & *** & ** \\
\bottomrule
\end{tabular}

EC: Enlarged Cardiomediastinum; Card: Cardiomegaly; AO: Airspace Opacity; LL: Lung Lesion; Edem: Edema; Cons: Consolidation; Pneu1: Pneumonia; Atel: Atelectasis; Pneu2: Pneumothorax; PE: Pleural Effusion; PO: Pleural Other; Frac: Fracture; SD: Support Devices

\label{tab:reschexpert}
\end{table*}

\begin{table*}[htbp]
\scriptsize
  \centering
    \caption{{The AUROC and AUPRC results of various models to classify the 13 classes (12 diseases and Support Devices) on MIMIC-CXR dataset. The upper part is the AUROC results and the lower part is the AUPRC results. *, ** and *** denote the significance levels of 0.1, 0.05 and 0.01, respectively. For each disease, the best results are bolded. The red text means our ImageGCN can perform better than the corresponding two baseline models. }
    }
\begin{tabular}{c|l|ccccccccccccc|c}
\toprule
\multicolumn{1}{l}{} &  & EC & Card & AO & LL & Edem & Cons & Pneu1 & Atel & Pneu2 & PE & PO & Frac & SD & mean \\
\midrule
 & A-GCN-PPS & {\color{red} 0.756} & {\color{red} 0.826} & {\color{red} 0.749} & {\color{red} 0.791} & {\color{red} 0.905} & {\color{red} 0.817} & {\color{red} 0.725} & {\color{red} 0.817} & {\color{red} 0.883} & {\color{red} 0.919} & {\color{red} 0.826} & 0.725 & {\color{red} 0.897} & {\color{red} 0.818} \\
 & A-GCN-APS & 0.752 & 0.816 & 0.743 & 0.768 & 0.887 & 0.801 & 0.723 & 0.804 & 0.877 & 0.901 & 0.826 & 0.764 & 0.879 & 0.811 \\
 & AlexNet & 0.745 & 0.819 & 0.738 & 0.721 & 0.901 & 0.805 & 0.703 & 0.812 & 0.850 & 0.911 & 0.791 & 0.666 & 0.889 & 0.796 \\
\cmidrule{2-16}
 & R-GCN-PPS & {\color{red} 0.764} & {\color{red} \textbf{0.828}} & 0.760 & {\color{red} \textbf{0.828}} & {\color{red} \textbf{0.909}} & {\color{red} 0.821} & {\color{red} \textbf{0.764}} & {\color{red} \textbf{0.826}} & 0.895 & {\color{red} \textbf{0.926}} & 0.849 & 0.761 & {\color{red} 0.907} & {\color{red} \textbf{0.834}} \\
 & R-GCN-APS & 0.760 & 0.820 & 0.761 & 0.778 & 0.892 & 0.818 & 0.744 & 0.817 & \textbf{0.897} & 0.909 & 0.853 & \textbf{0.825} & 0.890 & 0.828 \\
 & ResNet50 & 0.752 & 0.822 & 0.756 & 0.736 & 0.905 & 0.816 & 0.729 & 0.822 & 0.862 & 0.923 & 0.817 & 0.668 & 0.900 & 0.808 \\
\cmidrule{2-16}
 & V-GCN-PPS & {\color{red} \textbf{0.767}} & {\color{red} 0.826} & 0.760 & {\color{red} 0.826} & {\color{red} 0.908} & {\color{red} \textbf{0.826}} & {\color{red} 0.761} & {\color{red} 0.824} & {\color{red} \textbf{0.897}} & {\color{red} \textbf{0.926}} & 0.844 & 0.761 & {\color{red} \textbf{0.908}} & {\color{red} 0.833} \\
 & V-GCN-APS & 0.749 & 0.818 & \textbf{0.762} & 0.785 & 0.889 & 0.812 & 0.749 & 0.819 & 0.894 & 0.908 & \textbf{0.858} & 0.820 & 0.891 & 0.827 \\
 & VGGNet16BN & 0.743 & 0.818 & 0.748 & 0.715 & 0.904 & 0.816 & 0.717 & 0.821 & 0.856 & 0.920 & 0.796 & 0.678 & 0.900 & 0.802 \\
\cmidrule{2-16}\cmidrule{2-16}
 & p-val (RM-ANOVA) & ** & *** & * & *** & *** & ** & *** & *** & *** & *** & *** & *** & *** & *** \\
 & PPS \textgreater APS & * & *** &  & ** & *** & * & * & ** &  & *** &  &  & *** & *** \\
\multirow{-14}{*}{AUROC} & PPS \textgreater base & ** & *** & ** & *** & *** & ** & ** & *** & *** & ** & *** & *** & *** & *** \\
\bottomrule\toprule
 & A-GCN-PPS & {\color{red} 0.096} & {\color{red} 0.478} & {\color{red} 0.407} & {\color{red} 0.150} & {\color{red} 0.528} & {\color{red} 0.154} & {\color{red} 0.203} & 0.455 & {\color{red} 0.313} & {\color{red} 0.738} & 0.068 & 0.073 & {\color{red} 0.698} & {\color{red} 0.335} \\
 & A-GCN-APS & 0.094 & 0.461 & 0.398 & 0.122 & 0.480 & 0.145 & 0.191 & 0.436 & 0.313 & 0.690 & 0.087 & 0.099 & 0.659 & 0.321 \\
 & AlexNet & 0.087 & 0.465 & 0.397 & 0.085 & 0.510 & 0.138 & 0.182 & 0.456 & 0.228 & 0.726 & 0.044 & 0.041 & 0.687 & 0.311 \\
 \cmidrule{2-16}
 & R-GCN-PPS & {\color{red} \textbf{0.102}} & {\color{red} \textbf{0.479}} & {\color{red} 0.420} & {\color{red} \textbf{0.224}} & {\color{red} \textbf{0.548}} & {\color{red} \textbf{0.168}} & 0.233 & {\color{red} \textbf{0.476}} & 0.370 & {\color{red} \textbf{0.760}} & 0.081 & 0.102 & {\color{red} \textbf{0.730}} & {\color{red} \textbf{0.361}} \\
 & R-GCN-APS & 0.094 & 0.468 & 0.413 & 0.147 & 0.490 & 0.145 & \textbf{0.245} & 0.452 & \textbf{0.403} & 0.707 & \textbf{0.127} & \textbf{0.203} & 0.679 & 0.352 \\
 & ResNet50 & 0.092 & 0.473 & 0.417 & 0.100 & 0.535 & 0.155 & 0.213 & 0.472 & 0.288 & 0.755 & 0.062 & 0.044 & 0.713 & 0.332 \\
\cmidrule{2-16}
 & V-GCN-PPS & {\color{red} 0.101} & {\color{red} 0.475} & {\color{red} \textbf{0.421}} & {\color{red} 0.191} & {\color{red} 0.539} & {\color{red} 0.166} & 0.230 & 0.469 & 0.368 & {\color{red} 0.753} & 0.091 & 0.109 & {\color{red} 0.726} & {\color{red} 0.357} \\
 & V-GCN-APS & 0.098 & 0.470 & 0.419 & 0.149 & 0.489 & 0.156 & 0.233 & 0.458 & 0.376 & 0.703 & 0.118 & 0.178 & 0.679 & 0.348 \\
 & VGGNet16BN & 0.088 & 0.468 & 0.409 & 0.082 & 0.520 & 0.152 & 0.195 & 0.470 & 0.246 & 0.744 & 0.052 & 0.046 & 0.714 & 0.322 \\
\cmidrule{2-16}\cmidrule{2-16}
 & p-val (RM-ANOVA) & *** & ** &  & *** & *** & ** & ** & *** & *** & *** & *** & ** & *** & *** \\
 & PPS \textgreater APS & * & ** & ** & ** & *** & ** &  & ** &  & *** &  &  & *** & ** \\
\multirow{-14}{*}{AUPRC} & PPS \textgreater base & *** & ** & * & ** & *** & *** & ** &  & *** & ** & ** & ** & ** & *** \\
\bottomrule
\end{tabular}

EC: Enlarged Cardiomediastinum; Card: Cardiomegaly; AO: Airspace Opacity; LL: Lung Lesion; Edem: Edema; Cons: Consolidation; Pneu1: Pneumonia; Atel: Atelectasis; Pneu2: Pneumothorax; PE: Pleural Effusion; PO: Pleural Other; Frac: Fracture; SD: Support Devices

\label{tab:resmimiccxr}
\end{table*}

\subsection{Disease Identification} \label{sec:identifyresults}
{For the disease identification task, we use area under ROC curve (AUROC) {and area under the precision-recall curve (AUPRC)} to evaluate the performance of the models on three public datasets, \ie ChestX-ray14, CheXpert and MIMIC-CXR.}
\subsubsection{Results on ChestX-ray14}
Table \ref{tab:AUROC} {and \ref{tab:AUPRC}} show the AUROC {and AUPRC} scores of all the models on the 14 diseases on ChestX-ray14 dataset, respectively. From Table \ref{tab:AUROC} {and \ref{tab:AUPRC}}, with the same MPU type, GCN-PPS outperforms GCN-APS and the corresponding basic model in most of the diseases as well as in terms of the mean AUROC {and mean AUPRC} over the 14 diseases. V-GCN-PPS can even outperform the corresponding baselines V-GCN-APS and VGGNet16BN for all the 14 diseases {in terms of AUPRC}.

Table \ref{tab:AUROC} also lists some state-of-the-art AUROC results reported in the recent studies. Some studies like \cite{rajpurkar2017chexnet} that used a different training-validation-test split ratio or augmented the dataset are not listed. While no models can consistently achieve the best performance on all the 14 diseases, our V-GCN-PPS model achieves the best overall results among the 18 models. On 6 out of the 14 disease, V-GCN-PPS achieves the best AUROC scores. Our models V-GCN-PPS and R-GCN-PPS are the top 2 models in terms of the mean AUROC. We also calculate and list the mean ranks of the models for the 14 diseases in the last column of Table \ref{tab:AUROC}, where V-GCN-PPS and R-GCN-PPS achieve the top two minimum mean ranks among the 18 models. These results validate the efficacy of the proposed ImageGCN for CXR classification.

{We also conducted the repeated measures ANOVA {(RM-ANOVA)} and the post-hoc t-test to test the significance of performance differences among PPS, APS and the base models with 3 different backbones (\eg AlexNet, ResNet50 and VGGNet16BN) for each disease; the results are in the lower part of Table \ref{tab:AUROC} and \ref{tab:AUPRC}. From the results, in terms of AUROC, the performance differences are significant on all the 14 diseases except for Hern by {(RM-ANOVA) ($p=0.1$)}; PPS can outperform APS on 9 diseases at a significance level of 0.01 and on 13 diseases except for Hern at a significance level of 0.1; PPS can outperform base model on 8 diseases (\eg Card, Effu, Mass, Nodu, Pneu2, Emph, Fibr, and PT) at a significance level of 0.1. In terms of AUPRC, the performance differences are significant on 10 diseases except for Mass, Emph, Fibr, and Hern by {(RM-ANOVA) ($p=0.1$)}; PPS can significantly outperform APS and the base model on 10 diseases and 8 diseases at a significance level of 0.1, respectively.}


\subsubsection{Results on CheXpert}
We also validated our model on CheXpert and MIMIC-CXR datasets for disease identification. {(RM-ANOVA)} and the post-hoc tests are also conducted for each disease as well. {The results for CheXpert dataset are shown in Table \ref{tab:reschexpert}.} On CheXpert dataset, in terms of AUROC, the performance differences are significant on all the 13 classes by {(RM-ANOVA) ($p=0.1$)}; PPS can outperform APS on 5 classes at a significance level of 0.01 and on 9 classes at a significance level of 0.1; PPS can outperform base model on all the 13 classes at a significance level of 0.1. In terms of AUPRC, the performance differences are significant on 10 classes except for EC, Cons, and Atel by {(RM-ANOVA) ($p=0.1$)}; PPS can significantly outperform APS on 6 classes at a significance level of 0.05 and outperform the base model on 7 classes at a significance level of 0.1. In terms of mean value of both AUROC and AUPRC, PPS can outperform APS and the base model at a significance level of 0.05.

\subsubsection{Results on MIMIC-CXR}
{The results for MIMIC-CXR dataset are shown in Table \ref{tab:resmimiccxr}. }On MIMIC-CXR dataset, in terms of AUROC, the performance differences are significant on all the 13 classes at a significance level of 0.1 and on 10 classes at a significance level of 0.01 by {(RM-ANOVA)}; PPS can outperform APS on 4 classes at a significance level of 0.01 and on 9 classes at a significance level of 0.1; PPS can outperform the base model on all the 13 classes at a significance level of 0.05. In terms of AUPRC, the performance differences are significant on 12 classes except for AO by {(RM-ANOVA) ($p=0.05$)}; PPS can significantly outperform APS on 8 classes at a significance level of 0.1 and outperform the base model on 11 classes at a significance level of 0.05. In terms of mean AUROC {of the 13 classes}, PPS can outperform APS and the base model at a significance level of 0.01. In terms of mean AUPRC, PPS can outperform APS and the base model at a significance level of 0.05 and 0.01, respectively.

\subsubsection{Cross-dataset evaluation}
{To test the generalizability of our model, we also performed a cross-dataset evaluation for the models. Since CheXpert and MIMIC-CXR datasets have the same output classes, in the cross-dataset evaluation, we evaluate how the models trained on CheXpert training set perform on MIMIC-CXR test set. Table \ref{tab:chexpertonmimic} shows the results of cross-dataset evaluation for various models. In terms of AUROC, the performance differences are significant on 7 classes at a significance level of 0.1 by RM-ANOVA; but by a post-hoc paired t-test, PPS can outperform APS on 11 classes (except for EC and Pneu1) at a significance level of 0.1 and on 7 classes at a significance level of 0.01; PPS can outperform the base model on 11 classes at a significance level of 0.1. In terms of AUPRC, the performance differences are significant on 7 classes by RM-ANOVA at a significance level of 0.05; by a post-hoc paired t-test, PPS can outperform APS on 11 classes (except for EC and PO) at a significance level of 0.1 and outperform the base model on 10 classes (except for EC, Pneu1 and PO) at a significance level of 0.05. In terms of the overall mean AUROC and AUPRC over the 13 classes, PPS can outperform APS and the base model at a significance level of 0.01 and 0.1, respectively.
}

{Comparing Table \ref{tab:chexpertonmimic} and \ref{tab:reschexpert}, the same models trained on CheXpert training set can have quite different performances on MIMIC-CXR test set and CheXpert test set. This is caused by the discrepancy between the two datasets. Usually, due to the differences of data management and generation among different institutes, the data discrepancy in the same institute is smaller than that between different institutes, thus, models trained on CheXpert can have better performance on CheXpert dataset than that on MIMIC-CXR dataset. Similarly, due to the discrepancy, models trained on MIMIC-CXR training set can have better performance than the corresponding models trained on CheXpert training set when evaluated on MIMIC-CXR test set, by comparing Table \ref{tab:chexpertonmimic} and \ref{tab:resmimiccxr}.
}

\begin{table*}[htbp]
\scriptsize
  \centering
    \caption{{The results of cross-dataset evaluation. The AUROC and AUPRC results of various models trained on CheXpert dataset and evaluated on MIMIC-CXR dataset. The upper part is the AUROC results and the lower part is the AUPRC results. *, ** and *** denote the significance levels of 0.1, 0.05 and 0.01, respectively. For each disease, the best results are bolded. The red text means our ImageGCN can perform better than the corresponding two baseline models. }
    }
\begin{tabular}{c|l|ccccccccccccc|c}
\toprule
\multicolumn{1}{l}{} &  & EC & Card & AO & LL & Edem & Cons & Pneu1 & Atel & Pneu2 & PE & PO & Frac & SD & mean \\
\midrule
 & A-GCN-PPS & {\color{red} 0.567} & {\color{red} \textbf{0.770}} & {\color{red} 0.708} & {\color{red} 0.661} & {\color{red} 0.884} & {\color{red} 0.754} & 0.603 & {\color{red} \textbf{0.764}} & {\color{red} 0.752} & {\color{red} 0.895} & {\color{red} 0.701} & {\color{red} 0.610} & {\color{red} 0.839} & {\color{red} 0.731} \\
 & A-GCN-APS & 0.552 & 0.754 & 0.686 & 0.661 & 0.852 & 0.720 & 0.604 & 0.743 & 0.717 & 0.864 & 0.685 & 0.587 & 0.797 & 0.709 \\
  & AlexNet & 0.563 & 0.758 & 0.704 & 0.655 & 0.880 & 0.753 & 0.619 & 0.760 & 0.736 & 0.888 & 0.701 & 0.604 & 0.835 & 0.727 \\

\cmidrule{2-16}
& R-GCN-PPS & 0.530 & {\color{red} 0.768} & {\color{red} \textbf{0.721}} & {\color{red} \textbf{0.679}} & {\color{red} 0.885} & {\color{red} \textbf{0.772}} & {\color{red} \textbf{0.665}} & {\color{red} 0.761} & {\color{red} 0.771} & {\color{red} 0.902} & {\color{red} \textbf{0.757}} & {\color{red} 0.636} & {\color{red} \textbf{0.842}} & {\color{red} 0.745} \\
 & R-GCN-APS & 0.558 & 0.762 & 0.702 & 0.663 & 0.855 & 0.730 & 0.629 & 0.730 & 0.742 & 0.880 & 0.721 & 0.615 & 0.796 & 0.722 \\
 & ResNet50 & 0.527 & 0.713 & 0.671 & 0.609 & 0.842 & 0.690 & 0.606 & 0.702 & 0.700 & 0.857 & 0.655 & 0.573 & 0.781 & 0.687 \\
\cmidrule{2-16}
 & V-GCN-PPS & {\color{red} \textbf{0.576}} & {\color{red} 0.768} & {\color{red} \textbf{0.721}} & {\color{red} 0.673} & {\color{red} \textbf{0.887}} & {\color{red} 0.761} & {\color{red} 0.638} & {\color{red} 0.758} & {\color{red} \textbf{0.785}} & {\color{red} \textbf{0.904}} & {\color{red} 0.743} & {\color{red} \textbf{0.641}} & {\color{red} \textbf{0.842}} & {\color{red} \textbf{0.746}} \\
 & V-GCN-APS & 0.547 & 0.763 & 0.689 & 0.662 & 0.854 & 0.720 & 0.610 & 0.726 & 0.746 & 0.875 & 0.713 & 0.614 & 0.806 & 0.717 \\
 & VGGNet16BN & 0.520 & 0.709 & 0.690 & 0.634 & 0.854 & 0.687 & 0.596 & 0.701 & 0.718 & 0.878 & 0.672 & 0.582 & 0.809 & 0.696 \\
\cmidrule{2-16}
 & p-val (RM-ANOVA) &  & * &  & * & * &  &  & * & * & * &  &  & * &  \\
 & PPS \textgreater APS &  & * & ** & * & *** & *** &  & *** & *** & *** & ** & *** & *** & *** \\
\multirow{-14}{*}{AUROC} & PPS \textgreater base &  & * & * & * & * & * &  & * & * & * & * & * & * & * \\
\bottomrule \toprule
 & A-GCN-PPS & {\color{red} 0.042} & {\color{red} 0.414} & {\color{red} 0.353} & {\color{red} 0.064} & {\color{red} 0.459} & {\color{red} 0.110} & 0.123 & {\color{red} \textbf{0.373}} & {\color{red} 0.154} & {\color{red} 0.688} & 0.023 & {\color{red} 0.031} & {\color{red} \textbf{0.572}} & {\color{red} 0.262} \\
 & A-GCN-APS & 0.039 & 0.395 & 0.333 & 0.064 & 0.389 & 0.094 & 0.119 & 0.352 & 0.124 & 0.625 & 0.026 & 0.028 & 0.491 & 0.237 \\
 & AlexNet & 0.042 & 0.402 & 0.351 & 0.060 & 0.453 & 0.108 & 0.134 & 0.370 & 0.137 & 0.682 & 0.026 & 0.030 & 0.565 & 0.259 \\
\cmidrule{2-16}
 & R-GCN-PPS & 0.038 & {\color{red} \textbf{0.418}} & {\color{red} \textbf{0.373}} & {\color{red} \textbf{0.072}} & {\color{red} 0.467} & {\color{red} \textbf{0.112}} & {\color{red} \textbf{0.147}} & {\color{red} 0.366} & {\color{red} \textbf{0.208}} & {\color{red} \textbf{0.711}} & {\color{red} \textbf{0.034}} & {\color{red} \textbf{0.036}} & {\color{red} 0.554} & {\color{red} \textbf{0.272}} \\
 & R-GCN-APS & \textbf{0.045} & 0.405 & 0.347 & 0.068 & 0.407 & 0.099 & 0.130 & 0.339 & 0.168 & 0.660 & 0.029 & 0.033 & 0.492 & 0.248 \\
 & ResNet50 & 0.031 & 0.360 & 0.315 & 0.050 & 0.388 & 0.084 & 0.117 & 0.313 & 0.131 & 0.628 & 0.021 & 0.028 & 0.475 & 0.226 \\
\cmidrule{2-16}
 & V-GCN-PPS & {\color{red} 0.043} & {\color{red} 0.417} & {\color{red} 0.372} & {\color{red} 0.068} & {\color{red} \textbf{0.472}} & {\color{red} 0.110} & {\color{red} 0.142} & {\color{red} 0.370} & {\color{red} 0.201} & {\color{red} \textbf{0.711}} & {\color{red} 0.032} & {\color{red} \textbf{0.036}} & {\color{red} 0.558} & {\color{red} \textbf{0.272}} \\
 & V-GCN-APS & 0.037 & 0.404 & 0.336 & 0.063 & 0.406 & 0.092 & 0.123 & 0.338 & 0.170 & 0.647 & 0.030 & 0.033 & 0.507 & 0.245 \\
 & VGGNet16BN & 0.032 & 0.359 & 0.334 & 0.056 & 0.409 & 0.086 & 0.118 & 0.312 & 0.138 & 0.671 & 0.023 & 0.029 & 0.517 & 0.237 \\
\cmidrule{2-16}
 & p-val (RM-ANOVA) &  & * &  & * & * &  &  &  & ** & * &  & * & * &  \\
 & PPS \textgreater APS &  & *** & ** & * & *** & *** & * & *** & *** & *** &  & *** & *** & *** \\
\multirow{-14}{*}{AUPRC} & PPS \textgreater base &  & * & * & * & * & * &  & * & * & * &  & * & * & * \\
\bottomrule
\end{tabular}

EC: Enlarged Cardiomediastinum; Card: Cardiomegaly; AO: Airspace Opacity; LL: Lung Lesion; Edem: Edema; Cons: Consolidation; Pneu1: Pneumonia; Atel: Atelectasis; Pneu2: Pneumothorax; PE: Pleural Effusion; PO: Pleural Other; Frac: Fracture; SD: Support Devices

\label{tab:chexpertonmimic}
\end{table*}

\subsection{Ablation study}
We also conducted a multiple regression with a GLM model on the 4 types of features in the ChestX-ray14 dataset (\eg person, age, gender, view) to find which feature contributes most to the classification for each disease in a linear model. The p-values of the fitted model and the prediction results are shown in Table \ref{tab:pval}. In Table \ref{tab:pval}, most of the p-values are very small, which means that the features can correlate a lot with the corresponding disease. The poor prediction results (most of the AUROCs are around 0.5 and most of AUPRCs are around 0.1) indicate that only the 4 types of features are not enough for the disease prediction. In our work, these features are used as additional information to define the relation between images. From Table \ref{tab:pval}, 3 types of features (\eg person, age, gender) show somewhat large p-values for disease Consolidation, this may imply that involving these features cannot get much performance improvement for Consolidation identification. Similarly, 2 types of features (\eg person and gender) are likely not to contribute much to the Pneumonia identification. Our results in Table \ref{tab:AUROC} also validate the findings, where the PPS cannot get a significant improvement over the base model for diseases Consolidation and Pneumonia.

We have also conducted the ablation study to determine the importance of each relation for disease identification. The results of mean AUROC are listed in Table \ref{tab:ablation}, where each of the 4 types of relation information can improve the performance, and using all the relations can achieve the greatest improvements. When using a single relation, the \textit{person} relation performs better than the other 3 relations. This is as expected because in the ChestX-ray14 dataset the \textit{person} relation can imply the other 3 relations to a large extent. According to our statistics, for the 30,805 persons in the dataset, CXRs of the same person always have the same gender, 22,371 persons have their CXRs recorded at the same age,  23,682 persons's CXRs have only one view. We can also found that then \textit{age} relation contributes little to the overall disease identification results. The \textit{gender} relation contributes to the improvements more than the \textit{view} relation.

\begin{table*}[htbp]
\scriptsize
  \centering
    \caption{{The p-values of the 4 types of features contributing to each disease using a multiple regression with a GLM model and the prediction results in terms of AUROC and AUPRC. p-values greater than 0.05 are marked red.}
    }
\begin{tabular}{l|lllllllllllllll}
\toprule
   & Atel & Card & {Cons} & Edem &	Effu &	Emph &	Fibr &	Hern &	Infi &	Mass &	Nodu & PT &	Pneu1 &	Pneu2  \\
   \midrule
person & 1E-04 & 7E-05 & {\color{red} 1E+00} & 2E-03 & 1E-26 & 3E-02 & 7E-31 & 3E-02 & 4E-67 & 4E-11 & 9E-10 & 3E-02 & {\color{red} 8E-01} & 1E-29 \\
age & 3E-105 & {\color{red} 1E-01} & {\color{red} 7E-01} & 3E-02 & 9E-84 & 2E-18 & 1E-26 & 8E-33 & 3E-04 & 2E-14 & 2E-22 & 8E-24 & 9E-03 & {\color{red} 5E-01} \\
gender & 1E-07 & 7E-21 & {\color{red} 1E+00} & 1E-05 & 6E-04 & 5E-08 & {\color{red} 9E-01} & 3E-04 & {\color{red} 4E-01} & 4E-09 & 3E-02 & 1E-05 & {\color{red} 5E-01} & 3E-14 \\
view & 9E-92 & 1E-03 & 6E-243 & 0E+00 & 1E-125 & {\color{red} 2E-01} & 3E-54 & 3E-06 & 2E-255 & {\color{red} 8E-02} & 3E-16 & 3E-30 & 7E-24 & 2E-06 \\
\midrule
AUROC & 0.603 & 0.561 & 0.644 & 0.758 & 0.592 & 0.592 & 0.706 & 0.847 & 0.614 & 0.556 & 0.576 & 0.600 & 0.603 & 0.551 \\
AUPRC & 0.138 & 0.034 & 0.064 & 0.045 & 0.156 & 0.031 & 0.033 & 0.014 & 0.235 & 0.060 & 0.069 & 0.045 & 0.018 & 0.059 \\
\bottomrule
\end{tabular}
\label{tab:pval}
\end{table*}

\begin{table}[!t]
  \centering
  \caption{The mean AUROC results of V-GCN-PPS model using each of the 4 types of relations on ChestX-ray14 dataset. }
    \begin{tabular}{cccc|cc}
    \toprule
     {person} & {age} & {gender} &{view}  & {no relations} & all   \\
    \midrule
     0.829 & 0.824 & 0.826 & 0.825  & 0.824 & \textbf{0.832}\\
    \bottomrule
    \end{tabular}%
  \label{tab:ablation}%
\end{table}%

\begin{table}[!t]
  \centering
  \caption{The two parts of test set and the mean AUROC performances of V-GCN-PPS and the basic VGGNet16BN on the two parts. { Paired t-tests show the superiority of V-GCN-PPS over VGGNet16BN ($p=0.046$) and known part over unknown part ($p=0.021$).}}
    \begin{tabular}{l|r|r|r}
    \toprule
          & {known } & {unknown} & {overall} \\
    \midrule
    \#persons & 7,799 & 3,855 & 11,654 \\
    \#images & 18,336 & 4,088 & 22,424 \\
    \midrule
    V-GCN-PPS (all relations) & 0.827 & 0.787 & 0.832 \\
    VGGNet16BN (no relations) & 0.820 & 0.785 & 0.824 \\
    \bottomrule
    \end{tabular}%
  \label{tab:AUROCperson}%
\end{table}%

\begin{table*}[!t]
\footnotesize
  \centering
  \caption{The comparison results of disease localization. The best result in each cell is bolded. {The differences among the models are significant by RM-ANOVA for T(IoU)=0.1 ($p=0.008$) and T(IoU)=0.5 ($p=0.013$).}}
    \begin{tabular}{c|l|cccccccc}
    \toprule
    \multicolumn{1}{l|}{T(IoU)}  & \multicolumn{1}{c|}{model} & \multicolumn{1}{c}{Atelectasis} & \multicolumn{1}{c}{Cardiomegaly} & \multicolumn{1}{c}{Effusion} & \multicolumn{1}{c}{Infiltration} & \multicolumn{1}{c}{Mass} & \multicolumn{1}{c}{Nodule} & \multicolumn{1}{c}{Pneumonia} & \multicolumn{1}{c}{Pneumothorax} \\
    \midrule
    \multirow{3}{*}{0.1}  & A-GCN-PPS (ours) & \textbf{0.4889} & 0.9932 & {\textbf{0.6667}} & \textbf{0.6667} & {\textbf{0.4706}} & 0.0000 & {\textbf{0.6417}} & \textbf{0.3469} \\

          &       AlexNet & 0.3889 & \textbf{1.0000} & 0.6144 & 0.5285 & \textbf{0.4706} & \textbf{0.0253} & 0.5833 & 0.3265 \\
          &       A-GCN-APS & 0.3000 & 0.9863 & 0.5294 & 0.4634 & 0.2824 & 0.0127 & 0.5167 & 0.2755 \\
    \midrule
    \multirow{3}{*}{0.5}  & A-GCN-PPS (ours) & \textbf{0.0222} & {\textbf{0.3836}} & 0.0458 & {\textbf{0.1138}} & 0.0471 & 0.0000 & {\textbf{0.0750}} & {\textbf{0.0408}} \\
          &       AlexNet & 0.0111 & 0.2260 & \textbf{0.0784} & 0.0569 & \textbf{0.0824} & 0.0000 & \textbf{0.0750} & 0.0306 \\
          &       A-GCN-APS & 0.0000 & 0.3082 & 0.0327 & 0.0325 & 0.0235 & 0.0000 & 0.0500 & 0.0204 \\
    \bottomrule
    \end{tabular}  %

  \label{tab:bbox}%
\end{table*}%

As the \textit{person} relation is the most important for disease classification, we also conduct an experiment to compare the performances between images having and not having \textit{person} relations to the training set. In this experiments, we split the test set into 2 parts, known part and unknown part; known part consists of images having \textit{person} relations to the training set, \ie images of known persons that present in the training set, the unknown part consists of images of unknown persons. The summary information of the two parts and the performances of V-GCN-PPS with all relations and VGGNet16BN are listed in Table \ref{tab:AUROCperson}. {We conducted a paired t-test on the results between V-GCN-PPS and VGGNet16BN, the test results show that V-GCN-PPS can significantly outperform VGGNet16BN on different test sets ($p=0.046$).} In addition, in Table \ref{tab:AUROCperson}, both models perform better on known part than unknown part as expected ($p=0.021$), because the images of known persons would be more similar to the training set than those of unknown persons. For V-GCN-PPS, the explicit \textit{person} relation between test images and the training data can further contribute to the identification task, thus V-GCN-PPS achieves better performance on known part than VGGNet16BN. The performance gain of V-GCN-PPS on unknown part should be from the other 3 types of relations.

\subsection{Disease Localization}  \label{sec:bboxresults}
The ChestX-ray14 dataset provides 984 labelled pathology bounding box (Bbox) annotations for 880 CXR images by board-certified radiologists, which can be used as the ground truth of the disease localization task. The provided Bboxes correspond to 8 of the 14 diseases; some CXR images can have multiple Bbox annotations corresponding to different diseases.

In our experiments, we adopt the weakly supervised learning scheme \cite{oquab2015object} for disease localization, where no annotations are used for training. The provided Bbox annotations are used as ground truths to evaluate the performance of disease location of a model trained with image-level labels. For each image, we generate a heatmap normalized to $[0,255]$ with the MPU of self-connection in ImageGCN following the idea of CAM \cite{zhou2016learning}. We segment the heatmap by a threshold of 180, and generate Bboxes to cover the activated regions in the binary map. For a detect BBox and the ground truth (GT), we define IoU as intersection over union ratio, \ie
\begin{equation}
    \mbox{IoU} = \frac{Area(\mbox{BBox} \cap \mbox{GT})}{Area(\mbox{BBox} \cup \mbox{GT})}
\end{equation}
A correct localization is defined when the corresponding IoU is greater than a customized threshold T(IoU).

\begin{figure}[!t]
  \centering
  \subfloat[Atelectasis]{\includegraphics[width=.125\linewidth]{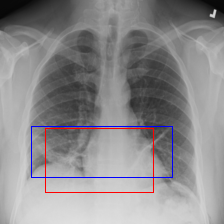}\includegraphics[width=.125\linewidth]{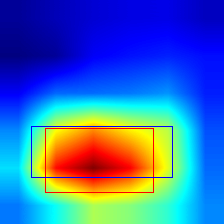}\includegraphics[width=.125\linewidth]{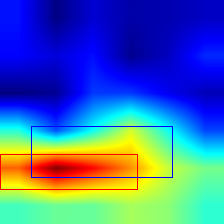}\includegraphics[width=.125\linewidth]{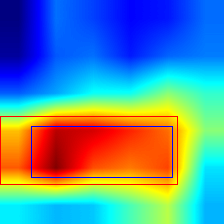} }
  \subfloat[Cardiomegaly]{ \includegraphics[width=.125\linewidth]{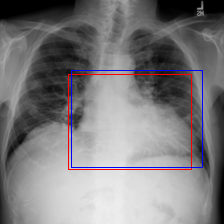}\includegraphics[width=.125\linewidth]{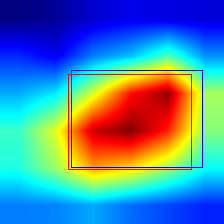}\includegraphics[width=.125\linewidth]{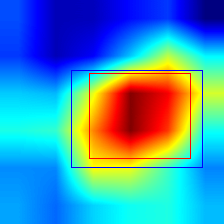}\includegraphics[width=.125\linewidth]{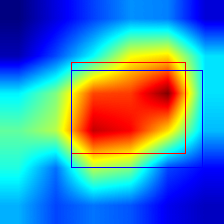} }

  \subfloat[Effusion]{\includegraphics[width=.125\linewidth]{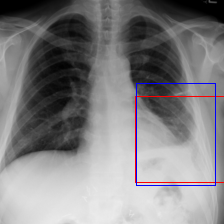}\includegraphics[width=.125\linewidth]{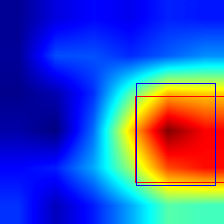}\includegraphics[width=.125\linewidth]{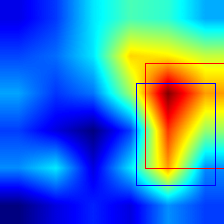}\includegraphics[width=.125\linewidth]{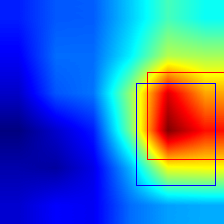} }
  \subfloat[Infiltratio]{ \includegraphics[width=.125\linewidth]{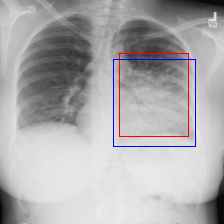}\includegraphics[width=.125\linewidth]{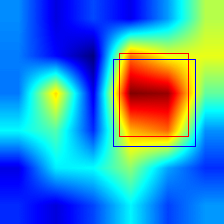}\includegraphics[width=.125\linewidth]{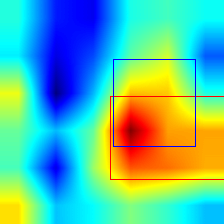}\includegraphics[width=.125\linewidth]{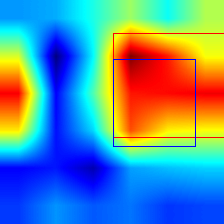} }

  \subfloat[Mass]{\includegraphics[width=.125\linewidth]{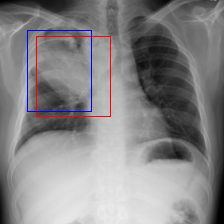}\includegraphics[width=.125\linewidth]{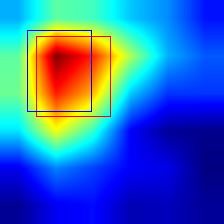}\includegraphics[width=.125\linewidth]{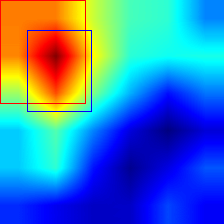}\includegraphics[width=.125\linewidth]{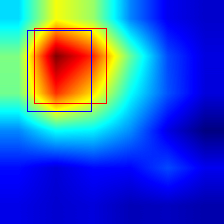} }
  \subfloat[Nodule]{ \includegraphics[width=.125\linewidth]{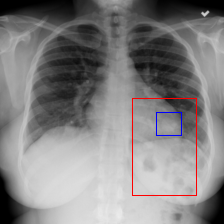}\includegraphics[width=.125\linewidth]{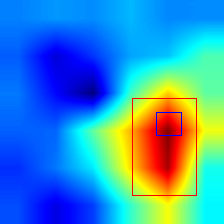}\includegraphics[width=.125\linewidth]{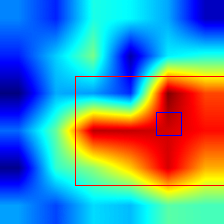}\includegraphics[width=.125\linewidth]{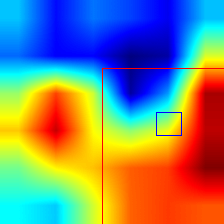} }

  \subfloat[Pneumonia]{\includegraphics[width=.125\linewidth]{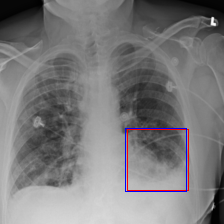}\includegraphics[width=.125\linewidth]{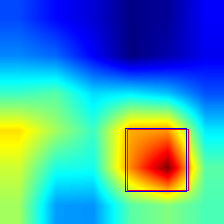}\includegraphics[width=.125\linewidth]{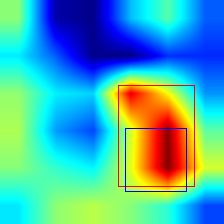}\includegraphics[width=.125\linewidth]{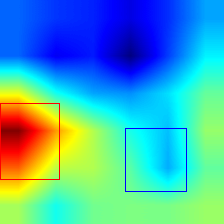} }
  \subfloat[Pneumothorax]{ \includegraphics[width=.125\linewidth]{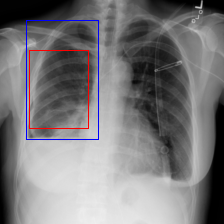}\includegraphics[width=.125\linewidth]{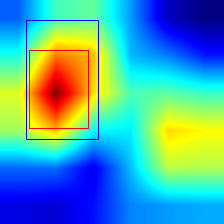}\includegraphics[width=.125\linewidth]{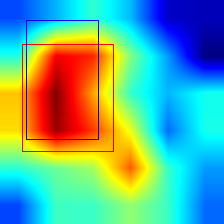}\includegraphics[width=.125\linewidth]{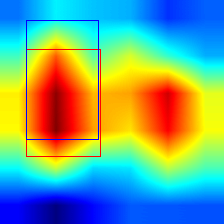} }

  \caption{The Bbox results of the models for the 8 diseases. For each cell, the first image is the original CXR with ground truth Bbox (blue) and Bbox generated by A-GCN-PPS (red). the second to fourth images are the generated heatmap and Bbox of A-GCN-PPS, A-GCN-APS and the base AlexNet, respectively. The blue Bbox is the ground truth, the red one is generated by the  respective models.     }
  \label{fig:boundingbox}
\end{figure}

The accuracy results of ImageGCN with AlexNet MPU (A-GCN-PPS) for disease localization are listed in Table \ref{tab:bbox}. From Table \ref{tab:bbox}, our ImageGCN A-GCN-PPS outperform the baselines in most cases. A-GCN-PPS performs best for 6 diseases if T(IoU)=0.1 and for 5 diseases if T(IoU)=0.5. {We conducted the {(RM-ANOVA)} and the post-hoc t-test on the results of disease localization for T(IoU)=0.1 and 0.5. The results of {(RM-ANOVA)} show significant performance differences among A-GCN-PPS, A-GCN-PPS, and AlexNet for both T(IoU)=0.1 ($p=0.008$) and 0.5 ($p=0.013$). The post-hoc t-tests show that A-GCN-PPS can significantly outperform A-GCN-APS ($p=0.003 $) and AlexNet ($p=0.036 $) for T(IoU)=0.1. For T(IoU)=0.5, A-GCN-PPS can significantly outperform A-GCN-APS ($p=0.008 $), but cannot significantly outperform AlexNet ($p=0.186 $).} Fig. \ref{fig:boundingbox} shows example localization qualitative results of A-GCN-PPS compared to the results of the baselines. From \ref{fig:boundingbox}, it can be seen that our ImageGCN with AlexNet MPU and PPS usually have smaller and more accurate Bboxes than the baselines.

\section{Discussion}
In the real world, relations can exist between images and the relations can contribute to the image representation. The proposed ImageGCN is able to model these image-level relations for image representation. To our best knowledge, this is the first study to model natural image-level relations for image representation learning. By the proposed idea of message passing unit, we extend GCN to graphs with high-dimensional or unstructured features. Also, by incorporating the ideas of GraphSAGE and R-GCN, ImageGCN can be trained by mini-batch inductively on multi-relational Graphs.

A special novelty of our work is the construction of the ImageGraph to incorporate the natural relations between images. Although many previous studies have applied GCN for image classification, their graph constructions are quite different from ours. The graphs that GCN was built on for image classification mainly involve 3 types in previous studies. (1) Label graph: the graph constructed with labels as nodes and label correlation dependencies as edges for multi-label learning, \eg \cite{chen2019multi}. Label graphs are usually small with size equal to the number of labels. (2) Knowledge graph: an external graph to describe the relations between entities where a entity usually represents a category for image classification, \eg \cite{wang2018zero}. (3) Similarity graph: an internal graph constructed based on the similarity between node features, \eg \cite{garcia2017few,yu2021resgnet,yu2021cgnet,hong2020graph}. Similarity graph cannot provide extra information {besides the internal features of the nodes} since the relations are derived from the features. The ImageGraph in our work is constructed with the original images as nodes and the natural relations between images as edges, thus it cannot be included in any of the above 3 types and can be the 4th type of graphs for image representation. Another advantage of ImageGCN is the end-to-end characteristic, that is, the model only need to be optimized one time under one single criterion. Previous studies like ResGNet-C \cite{yu2021resgnet} and CGNet \cite{yu2021cgnet} are a two-step training framework rather than an end-to-end framework, because they first train the CNN architecture to get the initial image representations, and then train a GCN to get the final image representations for classification.

After an ImageGCN model is trained, in the test process, if the training set is still available, for images in a test batch, we applied Algorithm \ref{al:subgraph} to get their representations. If the training set is no longer available, our ImageGCN model is also flexible for new images that have no relations with training samples, where only self-connection is taken into account (Eq. \ref{eq:spimagegcn}). In the extreme case when no relation exist in the ImageGraph, ImageGCN is decreased to a traditional deep learning framework.

ImageGCN is orthogonal to current CNN architectures. A CNN architecture can serve as a message passing unit in ImageGCN. In our experiments on ChestX-ray14, {CheXpert, and MIMIC-CXR datasets.} we tried AlexNet, ResNet50 and VGGNet16BN as the backbone of MPU, and reached a consistent conclusion. Moreover, by designing different MPUs, ImageGCN can be applied to different data modal. For example, we can use LSTM or RNN as the backbone of MPU to model the relation between texts, which would be one of our future research directions.

In our experiments on ChestX-ray14, {CheXpert, and MIMIC-CXR }datasets, using ImageGCN we incorporated 4 types of relations between CXR images into image representations for disease identification and localization. We explored how the relations contribute to the disease identification task and found that the \textit{person} relation is the most important among the 4 types of relations as expected. Although the \textit{person} relation is often unavailable for new patients in real-world scenarios, the other 3 types of relations can also improve the performance of identification. For known patients that have been recorded the CXR before, ImageGCN can reach even more gains for disease identification for new CXR images.

There may be two possible limitations in this study. (1) In the training process, besides the images in the batch, their sampled neighbors are also input to the GCN model for propagation, thus ImageGCN could cost more memories and more computation times than a base model without GCN. (2) The labels in the 3 CXR datasets (\eg ChestXray-14, CheXpert and MIMIC-CXR) are all generated from the radiological reports using natural language processing tools and are not 100\% accurate. This may cause some gaps between the classification results and the truth.

\section{Conclusion}
We propose ImageGCN to model relations between images and apply it to CXR images for disease identification and disease localization. ImageGCN extends the original GCN to high-dimensional or unstructured data, and incorporate the idea of R-GCN and GraphSAGE for batch propagation on multi-relational ImageGraphs. We also introduce the PPS scheme to reduce the complexity of ImageGCN. ImageGCN leads to better model consistency over related images, hence better explainability. The Experimental results on  {3 open-source x-ray datasets, ChestX-ray14, CheXpert and MIMIC-CXR} demonstrate that ImageGCN outperforms respective baselines in both disease identification and localization and can achieve comparable and often better results than the state-of-the-art methods.

\bibliographystyle{IEEEtran}
\bibliography{latex/tmi/egbib}

\end{document}